\documentclass[lettersize,journal]{IEEEtran}
\usepackage{amsmath,amsfonts}
\usepackage{bm} % Import bm package
\usepackage{algorithm}
\usepackage{algpseudocode} % for algorithm sudo code
\usepackage{array}
\usepackage[caption=false,font=normalsize,labelfont=normalsize,textfont=normalsize]{subfig}
\usepackage{textcomp}
\usepackage{stfloats}
\usepackage{url}
\usepackage{verbatim}
\usepackage{graphicx}
\usepackage[inkscapelatex=false]{svg}
\usepackage{cite}
\usepackage{tabulary}
\usepackage{mathtools}    
\usepackage{amsthm}
\newtheorem{remark}{Remark}

\usepackage{array}
\newcolumntype{P}[1]{>{\centering\arraybackslash}p{#1}}

\usepackage{url}
\usepackage[hidelinks]{hyperref}

\newtheorem{theorem}{Theorem}
\newtheorem{definition}{Definition}

\begin{document}
\title{Stream Function-Based Navigation for Complex Quadcopter Obstacle Avoidance}


%\markboth{Submitted to IEEE TRANSACTIONS ON SYSTEMS, MAN, AND CYBERNETICS: SYSTEMS}%

\author{
    \vskip 1em
    Sean Smith, \emph{Student Member, IEEE},
    Emmanuel Witrant,
    and Ya-Jun Pan, \emph{Senior Member, IEEE} \\
    \thanks{This work was supported by the Natural Sciences and Engineering Research Council (NSERC) of Canada, the France Canada Research Fund (FCRF), and the Conseil National des Universités.}
    \thanks{S. Smith and E. Witrant are with the GIPSA-lab, Université Grenoble Alpes - CNRS, F-38000, Grenoble, France, and the Department of Mechanical Engineering, Dalhousie University, Halifax NS B3H 4R2, Canada (e-mails: {\tt\small sean.smith@univ-grenoble-alpes.fr}, {\tt\small emmanuel.witrant@univ-grenoble-alpes.fr}).}
    \thanks{Ya-Jun Pan is with the Advanced Control and Mechatronics Laboratory, Department of Mechanical Engineering, Dalhousie University, Halifax NS B3H 4R2, Canada (e-mail: {\tt\small yajun.pan@dal.ca}).}
}




\maketitle

\begin{abstract}
This article presents a novel stream function-based navigational control system for obstacle avoidance, where obstacles are represented as two-dimensional (2D) rigid surfaces in inviscid, incompressible flows. The approach leverages the vortex panel method (VPM) and incorporates safety margins to control the stream function and flow properties around virtual surfaces, enabling navigation in complex, partially observed environments using real-time sensing. To address the limitations of the VPM in managing relative distance and avoiding rapidly accelerating obstacles at close proximity, the system integrates a model predictive controller (MPC) based on higher-order control barrier functions (HOCBF). This integration incorporates VPM trajectory generation, state estimation, and constraint handling into a receding-horizon optimization problem. The 2D rigid surfaces are enclosed using minimum bounding ellipses (MBEs), while an adaptive Kalman filter (AKF) captures and predicts obstacle dynamics, propagating these estimates into the MPC-HOCBF for rapid avoidance maneuvers. Evaluation is conducted using a PX4-powered Clover drone Gazebo simulator and real-time experiments involving a COEX Clover quadcopter equipped with a 360$^{\circ}$ LiDAR sensor.

\end{abstract}

\begin{IEEEkeywords}
Autonomous navigation, stream functions, model predictive control, higher-order control barrier function, quadcopter.
\end{IEEEkeywords}

% \fi
\IEEEpeerreviewmaketitle

\section{Introduction}

\IEEEPARstart{D}{espite} significant advancements in robotic systems, autonomous navigation in unknown and dynamic environments remains a challenging open problem. Limited sensor range, combined with the partial observability of the environment, introduces substantial difficulties in ensuring both safe operation and successful arrival at a designated destination.

In this article, we explore fluid dynamic theory for the real-time avoidance of complex and potentially highly dynamic obstacles. We propose a new algorithm for the detection, representation, and avoidance of partially observed obstacles with arbitrary shapes by modeling the detected surfaces as 2D surfaces in inviscid flow. By leveraging smooth velocity fields generated from stream functions, and combining potential flow theory with the panel method using the principle of superposition, the drone can avoid obstacles of varied shapes. Stream function-based navigation \cite{Stream_vessel,Stream_vessel2,Harmonic_potential,stream_functions} and the panel method \cite{vpm_drone_3D,vpm_online,vpm_online2,enhanced_VPM,pre_fake_robust} have been widely studied for avoiding arbitrary obstacles. However, the application of these methods in real-time scenarios remains challenging due to computational limitations and environmental uncertainties, necessitating further research and development.

%However, as the dynamic unpredictability of the obstacles increases, the reliability of a vortex panel method algorithm becomes limited in its avoidance capabilities alone. 
In this article, we present a novel approach that combines the VPM with optimal control to achieve robust, real-time dynamic obstacle avoidance. By leveraging stream function theory, partially observed obstacles are modeled as 2D surfaces using inviscid fluid dynamics. Inspired by rigid sail analysis \cite{jackson1984two}, such models enable the avoidance of obstacles with arbitrary shapes. The MPC framework tracks setpoints generated by the VPM, while incorporating HOCBF constraints \cite{HOCBF}. These constraints account for higher-order relative dynamics between the drone and obstacles, ensuring effective avoidance of highly dynamic obstacles in close proximity.

%Compared to existing work, our multi-layer dynamic obstacle avoidance algorithm exhibits high performance obstacle avoidance in the presence of accelerating obstacles, while simultaneously considering the challenges in obstacle state estimation based on partial observations.

This article's main contributions are listed as follows:
\begin{enumerate}
    \item A novel vortex panel method for real-time robot navigation is introduced, representing partially detected obstacles as 2D surfaces in inviscid flow to handle both concave and convex geometries. By introducing stream function control, using the Kutta condition \cite{kennedy}, and global convergence guarantees \cite{fahimi2009autonomous}, improvements are made over conventional potential methods, such as the artificial potential field (APF) method \cite{khatib1986real}, by avoiding local minima traps. 
    
   %Additionally, the proposed representation avoids the need for fitting obstacles with simple polygons, shapes...
    
    %An open-loop stream function control, using the Kutta condition \cite{kennedy}, and global convergence guarantees [??], are applied as flow-altering safety criterion. 

    %stream function control is paired with detected surface transformations
    
   % Coupling stream function control with transformations of the detected surface offers greater flexibility for navigation in proximity to complex obstacles.
    \item Unlike previous panel methods for drone navigation, which have been limited to offline scenarios and/or global knowledge of the environment \cite{vpm_drone_3D,pre_fake_robust,AIAA_1, AIAA_3}, the proposed approach enables real-time navigation in static and dynamic settings. To the authors' knowledge, this is the first application of the panel method to drones using real-time onboard sensing and navigation in moving environment applications.
    \item To address the limitations of the collision cone algorithm for panel method navigation at high relative speeds \cite{enhanced_VPM,modified_vpm}, and the lack of robustness in vortex-based fluid flow elements \cite{vpm_online}, an MPC-based HOCBF approach is integrated into the VPM framework. The MPC-HOCBF formulation incorporates higher-order obstacle dynamics, enabling the system to avoid rapidly accelerating obstacles in close proximity. Furthermore, it extends the 2D-VPM to 3D without introducing the computational and numerical complexities associated with 3D panel methods \cite{robotica_3D}. The 2D-VPM handles planar navigation, while the MPC-HOCBF makes necessary 3D adjustments.
    
    %collision cone-based source panel method (SPM) \cite{enhanced_VPM,modified_vpm} for dynamic obstacles, and the weak relative distance management of vortex-based flow elements \cite{vpm_online}, an MPC-based HOCBF approach is introduced to the VPM. The MPC-HOCBF framework incorporates higher-order obstacle dynamics to handle the avoidance of rapidly accelerating obstacles in close proximity, and extends the 2D-VPM to 3D without the computational and numerical complexities of 3D panel methods \cite{robotica_3D}. The 2D-VPM handles planar navigation, while the MPC-HOCBF makes necessary 3D adjustments.
    \item To integrate onboard sensing as feedback for the MPC-HOCBF, MBEs are used to parameterize detected rigid surfaces, serving as observations for a proposed AKF with an adaptive forgetting factor. Unlike previous contributions such as \cite{MPC_EKF_accel}, which rely on empirical relations to handle MBE fluctuation and consequently limit obstacle speed and shape, our AKF adaptively manages rapid oscillations in MBE feedback. This improves the prediction and detection of dynamic and complex-shaped obstacles.
    
    %, providing feedback for the MPC-HOCBF.
    %\item The MPC-HOCBF formulation explicitly incorporates higher-order obstacle dynamics, paired with high-relative degree control of the drone to enhance the performance in avoiding rapidly accelerating obstacles in close proximity.
\end{enumerate}
The proposed multi-layer control framework is evaluated through extensive simulations and experiments in the following scenarios: a) avoidance of static obstacles with complex shapes, including both convex and concave regions; b) avoidance of slow-moving dynamic obstacles; c) avoidance of high-acceleration, rotating dynamic obstacles; and d) 3D static and dynamic cases, demonstrating the control system's extendability to 3D with appropriate sensors.

\section{Literature Review}
This article builds on a large body of work in fluid dynamic-based obstacle avoidance, and MPC coupled with HOCBF for obstacle avoidance.

\subsection{Autonomous Navigation}
A wide variety of methods can be used for autonomous navigation, including Lyapunov methods \cite{Lyapunov_safe}, learning-based approaches \cite{Learning_avoidance}, vision-based methods \cite{vision_avoidance}, and sliding mode control (SMC) \cite{ricardo2023robust}. Among these methods, APFs have found extensive contributions for robot motion planning and control \cite{khatib1986real}. These methods model the environment as a potential field, using attractive and repulsive forces to guide the robot safely to its destination. However, potential functions can encounter challenges such as local minima traps or oscillations near obstacles. Stream functions and potential flows, as a subset of potential methods \cite{stream_functions, HPF_robotics}, mitigate these issues by avoiding local minima, with all local equilibria guaranteed to be saddle points \cite{kim1992real}.

%Stream functions \cite{Stream_vessel,Stream_vessel2,Harmonic_potential,stream_functions} yield

Stream functions yield collision-free trajectories of fluid particles, which can be adapted for robot navigation. While the stream function can be analytically defined for simple shapes like circles \cite{Harmonic_potential}, leading to effective circular obstacle representations \cite{Stream_vessel, Stream_vessel2}, this approach is limited when navigating unknown environments with complex convex and concave obstacles.

The panel method was introduced as a numerical method to evaluate flow around complex shapes. It was later combined with potential flows \cite{kim1992real} for robot navigation and control. It has since been adapted for 2D \cite{robotica_2D} and 3D \cite{robotica_3D} navigation, though these applications often represent the obstacles as simple convex polygons to reduce computation. In particular to drone systems, the panel method has been mainly limited to offline implementation \cite{vpm_drone_3D,pre_fake_robust} or a global knowledge of the environment \cite{AIAA_1,AIAA_3}. Experimental applications of the panel method in aerial vehicles appeared in \cite{vpm_drone_3D,pre_fake_robust,AIAA_3}, although these results were limited to offline settings, global knowledge of the environment, and static obstacles.

Online application of the VPM emerged in \cite{vpm_online} and \cite{vpm_online2}, using LiDAR sensors for a swarm of ground robots in static environments. Dynamic environments were considered in \cite{enhanced_VPM} and \cite{modified_vpm} using the source panel method (SPM) for ground robots, and paired with a collision cone approach in \cite{collision_cone}. This approach limits the method to slow moving obstacles. Open polygon shapes are considered in \cite{enhanced_VPM}, with safety maintained by controlling the velocity boundary conditions, however the versatility of the method is limited to a single safety parameter, and neither \cite{enhanced_VPM} nor \cite{modified_vpm} were experimentally validated. 

The above-mentioned approaches are limited in their ability to combine online path planning, tracking control, and constraint handling within a single unified framework. Consequently, they fail to address the discrepancy between the dynamic nature of the environments and the relatively slower response times of the higher-level planners. The following section investigates solutions to this problem.

\subsection{MPC and HOCBF}
%A large amount of research has explored using MPC for obstacle avoidance strategies \cite{manip_STO,NMPC_obsavoidance_mulTtraj,castillo2018model,zhao2021nonlinear,multiTraj_MPC_static_avoidance,ahn2022model,slack_perception,wang_MPCcbf_drone,multiDrone_mpc_cbfStatic,MPC_EKF_accel,Robust_MPC_HOCBF,MPC_kalman,MPC_EKF_gauss,MPC_GP,liu2024flexible}, due to its capacity to solve constrained optimization problems while simultaneously incorporating system dynamics, path planning, trajectory tracking, and avoidance constraints within a unified framework.

A large amount of research has explored using MPC for obstacle avoidance strategies \cite{manip_STO,NMPC_obsavoidance_mulTtraj,castillo2018model,zhao2021nonlinear,multiTraj_MPC_static_avoidance,slack_perception,wang_MPCcbf_drone,multiDrone_mpc_cbfStatic,MPC_EKF_accel,Robust_MPC_HOCBF,MPC_kalman,MPC_EKF_gauss,MPC_GP,liu2024flexible}, due to its capacity to solve constrained optimization problems while simultaneously incorporating system dynamics, path planning, trajectory tracking, and avoidance constraints within a unified framework.

The works in \cite{castillo2018model,slack_perception,multiTraj_MPC_static_avoidance} introduce the collision avoidance mechanism as a direct constraint for the MPC problem for static obstacles. Slack variables are used in  \cite{castillo2018model,slack_perception} to handle the violation of such constraints, which may occur in real systems. In \cite{manip_STO} and \cite{zhao2021nonlinear}, the avoidance is accomplished by integrating a potential function into the cost function; this also offers a method to avoid violating hard set constraints. 

Collision avoidance constraints have been approached through the use of control barrier functions (CBF) \cite{wang_MPCcbf_drone,multiDrone_mpc_cbfStatic,MPC_EKF_accel}, which effectively ensure the forward invariance of designated safe sets. HOCBFs \cite{HOCBF} extend CBFs by addressing constraints of arbitrary relative degrees, making them suitable for systems with higher-order dynamics. HOCBFs have been applied for collision avoidance \cite{liu2023high},\cite{HOCBF_drone}, and have been integrated with MPC control \cite{Robust_MPC_HOCBF}. However, despite these advancements, integrating MPC with HOCBF for dynamic obstacle avoidance in drones remains an open area of research.

%where CBFs have emerged as a powerful technique to ensure forward invariance of a designated safe set [??]. However, mainly static obstacles are considered in the mentioned cases, without need to propagate the obstacle dynamics within the prediction horizon.

Most of the above-mentioned works focus on the avoidance of static obstacles, and do not consider propagating the obstacles dynamics through the prediction horizon. Dynamic collision avoidance was investigated using MPC combined with a model-based Kalman filter in \cite{MPC_kalman} and \cite{MPC_EKF_gauss}, which predicted obstacle trajectories based on a constant velocity model. This method was compared to MPC with a model-free Gaussian Process learning-based approach \cite{MPC_GP}. However, the method in \cite{MPC_GP} relies on a simple Euclidean norm model that only becomes active when the robot is near the constraint boundary, potentially compromising safety and/or leading to constraint violations. 

Dynamic obstacle avoidance for robotic manipulators, using robust observers for state estimation was coupled with a MPC-CBF approach \cite{liu2024flexible}, and MPC with a potential function approach \cite{manip_STO}. These prior works address obstacles moving at a constant velocity while using lower order observers for state estimation, neglecting the higher-order obstacle dynamics. 

The work in \cite{NMPC_obsavoidance_mulTtraj} couples acceleration models with nonlinear MPC, however, the trajectory is limited to projectile motion, and the safety criteria relies on a simple Euclidean norm. In \cite{MPC_EKF_accel}, a Kalman filter using linear acceleration models is combined with MPC for ground robots. However, the variation in obstacle parametrization is addressed using an empirically determined correlation, which may not be extendable to a wide range of obstacle sizes and motions.

%which may not extend well to obstacles with varying sizes and motions.

%which may not be extendable to a wide range of obstacle sizes and motions.

In light of the mentioned works, we developed an MPC-HOCBF formulation combined with an AKF, which incorporates an adaptive mechanism to handle undesirable rapid variations in the observations and to identify obstacle dynamics online. The proposed VPM-MPC-HOCBF-AKF combines the online path planning from the VPM, tracking control, state estimation, and constraints handling into a single framework.

\section{Real-time Vortex Panel Method for Robot Navigation
}
\subsection{Background on Stream Functions}
Let $\Phi \subset \mathbb{R}^2$ be a 2D domain representing the flow region. The stream function $\psi: \Phi \times [0,T]\rightarrow \mathbb{R}$ is a scalar function used to describe incompressible flow, satisfying continuity \(\text{\(\nabla \cdot \boldsymbol{\nu} = 0\)}\) and the condition $\boldsymbol{\nu}\cdot \nabla \psi=0$, where $\boldsymbol{\nu} = [v_x^r, \ v_y^r]^T$ is the fluid velocity vector, and the flow may be viscous and/or rotational. The vorticity $\omega$ of the flow is defined as
\begin{equation}
    \omega = \nabla \times \boldsymbol{\nu} = - \nabla^2 \psi,
\end{equation}
where $\nabla = [\frac{\partial}{\partial x}, \ \frac{\partial}{\partial y}]^T$ is the 2D gradient operator. For irrotational flow, the stream function satisfies Laplace's equation $\nabla^2  \psi = 0$.

A stream function $\psi$ has a gradient perpendicular to the velocity field, this gives the velocity components $v_x^r$ and $v_y^r$ of the fluid
\begin{equation}\label{eq:velocity_dif}
 v_x^r = \frac{\partial\psi}{\partial y}, \ \ v_y^r = 
     -\frac{\partial\psi}{\partial x},
\end{equation}
which will later be used for robot collision-free reference tracking. A streamline is the line along which a stream function is constant $\psi(x,y) = C$, for some constant $C$. Physically, a streamline represents the trajectory of a fluid particle as it moves within the flow field.

The primary types of flow elements to consider are uniform flow, source, sink, and vortex flow. These elementary flow types can be superimposed to form more complex flow patterns. The stream function for a point vortex of strength $\zeta_o$ located at $(x_o, y_o)$, is given by
\begin{equation}
    \psi_o = -\frac{\zeta_o}{2\pi}\ln{\ell},
\end{equation}
where $\ell = \sqrt{(x-x_o)^2 + (y-y_o)^2}$ is the radial distance from the vortex center. 

For a source located at $(x_{w},y_w)$ and a sink (goal) at $(x_{g},y_g)$ with strengths $\zeta_{w}$ and  $\zeta_g$, respectively, the stream function is given by
\begin{equation} \label{eq:source_sink}
    \psi_{w,g} = \frac{\zeta_{w,g}}{2\pi}\tan^{-1}\left(\frac{y-y_{w,g}}{x-x_{w,g}}\right),
\end{equation}
where $\zeta_w > 0$ denotes a source and $\zeta_g < 0$ denotes a sink. The uniform flow stream function is simply given by $\psi_{\infty} = Q_{\infty}(y_i\cos{\vartheta_{\infty}} - x_i\sin{\vartheta_{\infty}})$ where $Q_{\infty}$ is the flow velocity magnitude and $\vartheta_{\infty}$ is the global angle of attack, which is given by
\begin{equation}
    \vartheta_{\infty} = \tan^{-1}{\left(\frac{y_g - y_w}{x_g - x_w}\right)}.
\end{equation}

\subsection{Vortex Panel Method Algorithm}\label{sec:VPM}

\begin{figure}[tb]
\centerline{\includegraphics[width=0.95\columnwidth]{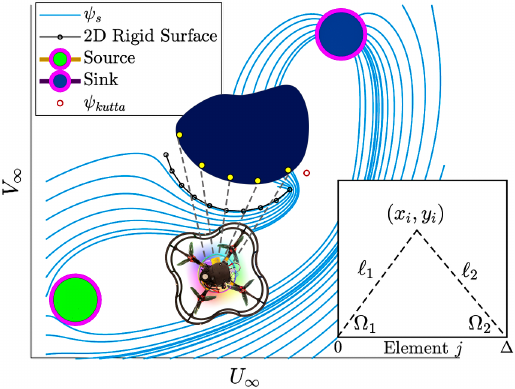}}
\caption{Structural overview of the vortex panel method (VPM) in a 2D environment, illustrating navigation from a source to a sink with the integration of rigid surfaces for maneuvering in unknown environments. The geometric relationship between a control point $i$ and a corresponding element $j$ is highlighted.}
\label{fig:nav_overlay}
\end{figure}

The insertion of a rigid body into a fluid flow field introduces a boundary condition requiring that the flow be tangent to the surface $\mathcal{G}$, implying that the normal component of the velocity is zero, making the solid surface a streamline of the flow. This condition is illustrated with the flow around the virtual 2D rigid surface in Fig. \ref{fig:nav_overlay}. Given the stream functions $\psi_{s}$, defined on the surfaces $\mathcal{G}(s)$ for $s \in \{1,...,M\}$, the boundary condition for $\nabla^2  \psi = 0$ can be expressed as
\begin{equation}\label{eq:boundary_condition}
    \psi = \psi_s, \ \boldsymbol{\nu} \cdot \mathbf{n} = 0 \ on \ \mathcal{G}(s), \ s \in \{1,...,M\},
\end{equation}
where $\mathbf{n}$ is the unit vector normal to the surface $\mathcal{G}(s)$.

A general point along the surface $\mathcal{G}$ of a rigid body, designated by $\mathcal{G}'$, has a vortex density of $\gamma(\mathcal{G}')$, which can be distributed along the surface. Assume that the rigid bodies surface of arbitrary shape is discretized into $N$ elements, where each element on the surface has a control point $C_i$ located at $(x_i,y_i)$ and each element has a vorticity density of $\gamma(\mathcal{G}_j)$. Integrating this contribution over the entire surface $\mathcal{G}$, resulting in $N$ integrals over the $N$ elements, and evaluating at control point $C_i$ gives the stream function along the surface as
\begin{equation}\label{eq:vorticity_distribution}
    \psi_s = -\frac{1}{2\pi}\sum_{j=1}^{N}\int_{\mathcal{G}_j}\gamma(\mathcal{G}_j)\ln{\ell(C_i,\mathcal{G}_j)} \,d\mathcal{G}_j.
\end{equation}
Following the steps in \cite{kennedy}, we begin by assuming that the boundary elements are straight lines of length $\Delta_j$, with control points located at their midpoints. The vorticity distribution $\gamma(\mathcal{G}')$ is assumed to be constant over each element, and denote this constant by $\gamma_j$. Applying the boundary condition \eqref{eq:boundary_condition} to the combined flow field, which consists of the uniform stream, the vorticity distribution from \eqref{eq:vorticity_distribution}, and the source and sink contributions from \eqref{eq:source_sink}, yields
%and assuming the elements are straight lines of length $\Delta_j$ with control points at their midpoints, and that $\gamma(S')$ is constant (denoted $\gamma_j$) over each element, applying the boundary condition \eqref{eq:boundary_condition} to the combined flow, which includes the uniform stream, the vorticity distribution from \eqref{eq:vorticity_distribution}, a source, and a sink from \eqref{eq:source_sink}, yields 

\begin{equation}\label{eq:stream_function_panel}
\begin{cases}
    \psi_s + \sum_{j=1}^{N} \gamma_jI_{ij} = \underbrace{\psi_{\infty} + \psi_{w} + \psi_{g}}_{\substack{F_i}},\\
    I_{ij} = -\frac{1}{2\pi}\left(x_i\ln{\frac{\ell_2}{\ell_1}} - \Delta_j \ln{\ell_2} + y_i(\Omega_1 - \Omega_2)\right).
    \end{cases}
\end{equation}
Term $I_{ij}$, $i \in \{1,...,N\}$, represents the geometric influence function of element $j$ on control point $i$, $\Omega_1 = \tan^{-1}(\frac{y_i}{x_i})$, $\Omega_2 = \tan^{-1}(\frac{y_i}{x_i - \Delta_j})$, $\ell_1 = \sqrt{x_i^2 + y_i^2}$, and $\ell_2 = \sqrt{(x_i-\Delta_j)^2 + y_i^2}$. Here, $(x_i,y_i)$ denotes the $i$th control point in the reference frame of the current $j$th panel. The potential flow within the domain $\Phi \subset \mathbb{R}^2$ is determined by evaluating \eqref{eq:stream_function_panel} at the $N$ midpoints, leading to the linear system
\begin{equation}
     \sum_{j=1}^{N} \gamma_jI_{ij}=-\psi_s + F_i.
\end{equation}
Assuming the matrix $I_{ij}$ is invertible, which holds provided that panels do not overlap, define $K_{ij} = I_{ij}^{-1}$ giving the solution
\begin{equation}\label{eq:solve_gamma}
    \gamma_i = \sum_{j=1}^{N} K_{ij}(-\psi_s + F_j), \ \ i \in \{1,...,N\}.
\end{equation}

From Eq. \eqref{eq:solve_gamma}, there are $N+1$ unknowns: the $N$ vorticity strengths $\gamma_i$ and the obstacle surface stream function $\psi_s$, while there are only $N$ equations. To resolve this underdetermined system and solve for $\psi_s$, one additional equation must be introduced, which is addressed in the following subsection.

\subsection{Adaptation to Robot Navigation}

\subsubsection{VPM-A Kutta Condition for Safe Navigation}

To ensure the flow streamline detaches smoothly from the trailing edge of the detected surface and guides the quadcopter along a safe trajectory of fluid flow, a Kutta condition is applied. In this approach, $\psi_s$ is calculated for a point extended beyond the trailing edge (see Fig. \ref{fig:nav_overlay}), as outlined in  \cite{kennedy}, resulting in $N+1$ equations for the unknowns $\gamma_i$,  $i\in \{1,...,N\}$, and $\psi_{kutta}$. While the Kutta condition traditionally ensures smooth airflow detachment in aerodynamic applications, here it is leveraged for safe quadcopter navigation. The algorithm exploits the flow-altering properties of the Kutta condition as a safety mechanism, directing the flow field away from obstacles.

In this algorithm, the virtual surface is positioned with an offset from the detected obstacle (see Fig. \ref{fig:nav_overlay}), acting as a virtual barrier through the boundary condition \eqref{eq:boundary_condition}. The surface is maintained between the drone and the obstacle, while the Kutta condition $\psi_{kutta}$ directs the fluid flow, ultimately guiding the drone safely around the obstacle. The transformation process is detailed in Algorithm \ref{alg:VPM_A_kutta} for a single surface $\mathcal{G}$.

%, with input variables consisting of the global LiDAR readings of the detected surface $(\mathbf{x}_{\mathcal{G}}, \mathbf{y}_\mathcal{G})$, the detected surface transformation parameters $(\mu,\kappa,\ell_{kutta})$, and the 2D global position of the quadcopter $(p_x, p_y)$.

\begin{algorithm}[t]
\caption{Transformation of the Virtual Surface for the VPM-A Algorithm}
\label{alg:VPM_A_kutta}
\textbf{Input:} Global surface coordinates $\bm{Z}_k = [\mathbf{x}_{\mathcal{G}}, \mathbf{y}_\mathcal{G}]$, surface transformation parameters ($\mu$, $\kappa$, $\ell_{kutta}$), and the 2D global position of the quadcopter $(p_x, p_y)$.\\
\textbf{Output:} Transformed surface $\bm{\Tilde{Z}}_k^T$, and Kutta location $\bm{Z}_{kutta}$.
\begin{algorithmic}[1]
\State Compute the surface centroid: 
\vspace{1mm}
\Statex $x_c = \frac{1}{n}\sum_{i=1}^n \mathbf{x}_{\mathcal{G}_i}$, $\ y_c = \frac{1}{n}\sum_{i=1}^n \mathbf{y}_{\mathcal{G}_i}$
\vspace{1mm}
\State Compute minimum distance from the robot to the surface: 
%\vspace{0.1mm}
\Statex $d_{min} = \min \left\{ \sqrt{ \left( \mathbf{x}_{\mathcal{G}} - p_x \right)^2 + \left( \mathbf{y}_{\mathcal{G}} -p_y \right)^2 } \right\}$
\vspace{1mm}
%\State Compute shift: $d_{{shift}} = \mu \cdot d_{min}, \ 0 \leq \mu < 1$
\State Compute unit vector from the centroid to the robot: 
\vspace{1mm}
\Statex $\vec{\bm{q}}_{cr} = \begin{bmatrix} p_x \\ p_y \end{bmatrix} - \begin{bmatrix} x_c \\ y_c \end{bmatrix}$, $\ \hat{\vec{\bm{q}}}_{cr} = \frac{\vec{\bm{q}}_{cr}}{ \|\vec{\bm{q}}_{cr}\|}$
\vspace{1mm}
\State Shift the surface by $d_{{shift}} = \mu \cdot d_{min}, \ 0 \leq \mu < 1$: 
\vspace{1mm}
\Statex$\vec{\bm{s}} = d_{shift} \cdot \hat{\vec{\bm{q}}}_{cr}$, $\ \Tilde{\bm{Z}}_k^T = [\Tilde{\mathbf{x}}_{\mathcal{G}}, \tilde{\mathbf{y}}_{\mathcal{G}}]=\bm{Z}_k + \vec{\bm{s}}$
\vspace{1mm}
\State Compute the trailing edge unit vector, with $(\Tilde{\mathbf{x}}_{\mathcal{G}}, \Tilde{\mathbf{y}}_{\mathcal{G}})$ ordered such that the coordinates $(\Tilde{\mathbf{x}}_{\mathcal{G}_n}, \Tilde{\mathbf{y}}_{\mathcal{G}_n})$ are last in the direction of flow:
\vspace{1mm}
\Statex$\vec{\bm{q}}_{kutta} = \begin{bmatrix} \Tilde{\mathbf{x}}_{\mathcal{G}_{n}} - \Tilde{\mathbf{x}}_{\mathcal{G}_{n-1}} \\ \Tilde{\mathbf{y}}_{\mathcal{G}_{n}} - \Tilde{\mathbf{y}}_{\mathcal{G}_{n-1}} \end{bmatrix}$, $\ \hat{\vec{\bm{q}}}_{kutta} = \frac{\vec{\bm{q}}_{kutta}}{ \|\vec{\bm{q}}_{kutta}\|}$
\vspace{1mm}
\State Rotate the unit vector direction:
\vspace{1mm}
\Statex $\hat{\vec{\bm{e}}}_{kutta} = \begin{bmatrix} \cos(\kappa) & -\sin(\kappa) \\ \sin(\kappa) & \cos(\kappa) \end{bmatrix} \cdot \hat{\vec{\bm{q}}}_{kutta}$
\vspace{1mm}
\State Compute the Kutta location: 
\vspace{1mm}
\Statex $\bm{Z}_{kutta} = \begin{bmatrix} \Tilde{\mathbf{x}}_{\mathcal{G}_n} \\ \Tilde{\mathbf{y}}_{\mathcal{G}_n} \end{bmatrix} + \ell_{kutta} \cdot \hat{\vec{\bm{e}}}_{kutta}$
\vspace{1mm}
\State \textbf{return} $\bm{\Tilde{Z}}_k^T$, $\bm{Z}_{kutta}$
\end{algorithmic}
\end{algorithm}

\subsubsection[VPM-B Global Convergence Method]{VPM-B Global Convergence Method \cite{fahimi2009autonomous}}

%Inspired by the SPM in \cite{fahimi2009autonomous}, 

The following method automatically adjusts $\psi_s$, and subsequently the vortex strengths $\gamma_i$, to ensure that the sink location is a global minimum. To achieve this, consider the following inequality
\begin{equation}\label{eq:convergence}
    |\zeta_g| > \zeta_s > -|\zeta_g|,
\end{equation}
where $|\zeta_g|$ is the strength of the sink and
\begin{equation}\label{eq:total_strength}
    \zeta_s = \sum_{i=1}^{N}\gamma_i\Delta_i,
\end{equation}
is the total vortex strength of the surface $\mathcal{G}(s)$, where $\Delta_i$ is the length of each panel. 

%Using \eqref{eq:stream_function_panel} we can write $\sum_{j=1}^{N} \gamma_jI_{ij} = -\psi_k + R_i$. If $K_{ij} = I_{ij}^{-1}$, which is invertible as long as panels do not overlap, then we can solve for
%\begin{equation}\label{eq:solve_gamma}
%    \gamma_i = \sum_{j=1}^{N} K_{ij}(-\psi_k + R_j).
%\end{equation}

Combining \eqref{eq:solve_gamma}, \eqref{eq:convergence}, and \eqref{eq:total_strength}  gives the following bounds for the stream function
\begin{equation}\label{eq:psi_bounds}
    \begin{cases}
        \psi_s^{\min} < \psi_s < \psi_s^{\max}, \\ 
        \psi_s^{\max} = \frac{|\zeta_g| + \sum_{i=1}^{N}\Delta_i \sum_{j=1}^{N}K_{ij}F_j}{\sum_{i=1}^{N}\Delta_i\sum_{j=1}^{N}K_{ij}},\\
        \psi_s^{\min} = \frac{-|\zeta_g| + \sum_{i=1}^{N}\Delta_i \sum_{j=1}^{N}K_{ij}F_j}{\sum_{i=1}^{N}\Delta_i\sum_{j=1}^{N}K_{ij}},\\
        \psi_s^{0} = \frac{\sum_{i=1}^{N}\Delta_i \sum_{j=1}^{N}K_{ij}F_j}{\sum_{i=1}^{N}\Delta_i\sum_{j=1}^{N}K_{ij}}.
    \end{cases}
\end{equation}
Here, the stream function values $(\psi_s^{\min}, \psi_s^{0}, \psi_s^{{\max}})$ correspond to the vortex strengths $(|\zeta_g|, 0, -|\zeta_g|)$. A positive obstacle vortex strength $\zeta_s$ indicates that the flow tends to circulate counterclockwise (CCW) around the obstacle. Selecting $\psi_s$ outside the bounds $[\psi_s^{\min}, \psi_s^{\max}]$ may result in a velocity field that prevents the fluid particles from reaching the goal, instead trapping them in a circulating flow around the obstacle.

%If the obstacle vortex strength $\zeta_s$ is positive, it indicates that the flow will tend to travel counterclockwise (CCW) around the obstacle. Selecting $\psi_s$ outside the bounds $[\psi_s^{\min}, \psi_s^{\max}]$ may lead to a velocity field created by the vortex obstacle that prevents the fluid particles from reaching the goal location, instead causing them to remain in a flow that circulates around the obstacle.

Using Eq. \eqref{eq:psi_bounds}, the stream function is calculated as
\begin{equation}
\begin{cases}
    \psi_s = \psi_s^0 + sign(\xi)\left(\frac{|\xi|(\psi_s^{\max} - \psi_s^{\min})}{2} \right),\\
    \xi\in(-1,1),
    \end{cases}
\end{equation}
where $\xi$ is a user-defined parameter. For $\xi>0$, increasing $\xi$ causes the stream function $\psi_s$ to approach $\psi_s^{\max}$, while $\zeta_s$ approaches $-|\zeta_g|$, resulting in a clockwise (CW) flow around the obstacle. A larger $\xi$ increases the angular velocity of the fluid particles, which tends to draw the flow (and thus the robot) closer to the obstacle, particularly for concave obstacles. Conversely, for $\xi<0$, decreasing $\xi$ causes $\psi_s$ to approach $\psi_s^{\min}$, while $\zeta_s$ approaches $|\zeta_g|$, resulting in a CCW flow around the obstacle.

For the simulations presented in this work, the parameter $\xi$ is selected such that the Clover quadcopter navigates to the left when an obstacle is detected in the right field of view (FOV), and to the right if an obstacle is detected in the left FOV.

% \begin{algorithm}\label{alg:VPM_B}
% \caption{Vortex-Panel-Method Algorithm B}
% \begin{algorithmic}[1]
% \Function{VPM-B}{$I_{ij}, L_i, R_j, \Gamma_{g}, \xi$}
% \State $\psi_k^{\max}$ \eqref{eq:psi_max}
% \State $\psi_k$ \eqref{eq:psi_max}
% \State $\gamma_i$ \eqref{eq:solve_gamma}
% \Return $\gamma_i$
% \end{algorithmic}
% \end{algorithm}

Upon solving the system of equations \eqref{eq:solve_gamma} for the $N$ vortex strengths of each panel, the flow velocity at any point in the spatial domain $\Phi \subset \mathbb{R}^2$ can be evaluated using \eqref{eq:velocity_dif}. The velocity at point \emph{i}, induced by the vorticity of each panel $\gamma_j$ (see Fig. \ref{fig:nav_overlay}), can be determined from \eqref{eq:velocity_dif} and \eqref{eq:stream_function_panel}, which gives
\begin{equation}\label{eq:panel_velocity}
    \begin{bmatrix}
        v_{\hat{n}}^{r_i} \\ v_{\hat{\tau}}^{r_i}
    \end{bmatrix} = \frac{\gamma_j}{2\pi}
    \begin{bmatrix}
        \Omega_1 - \Omega_2 \\
        \ln{\frac{\ell_1}{\ell_2}} 
    \end{bmatrix}.
\end{equation}
Terms $v^{r_i}_{\hat{n}}$ and $v_{\hat{\tau}}^{r_i}$ are the normal and tangential velocity contributions from the vortex element \emph{j}, as observed in the reference frame of element \emph{j}. These contributions must be transformed into the global reference frame. Similarly, the velocity contribution at point $(x_i,y_i)$ in space, due to a source or sink of strength $\zeta_{w,g}$ at $(x_{w,g},y_{w,g})$ is determined using \eqref{eq:velocity_dif} and \eqref{eq:source_sink} giving

\begin{equation}\label{eq:source_vel}
    \begin{bmatrix}
        v^{r_i}_x \\ v^{r_i}_y
    \end{bmatrix} = \frac{\Gamma_{w,g}}{2\pi}
    \begin{bmatrix}
        \frac{x-x_{w,g}}{(x-x_{w,g})^2 + (y-y_{w,g})^2} \\
        \frac{y-y_{w,g}}{(x-x_{w,g})^2 + (y-y_{w,g})^2} 
    \end{bmatrix}.
\end{equation}
Uniform flow components $(v^{r_i}_{x,\infty}, v^{r_i}_{y,\infty})$ can be included to direct the flow distribution, where $v^{r_i}_{x,\infty} = Q_{\infty}\cos{\vartheta_{\infty}}$ and $v^{r_i}_{y,\infty} = Q_{\infty}\sin{\vartheta_{\infty}}$.

\begin{figure}[tb]
\centerline{\includegraphics[width=\columnwidth]{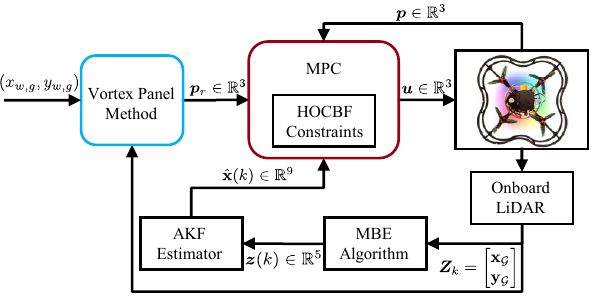}}
\caption{Control architecture for the proposed VPM-MPC-HOCBF-AKF algorithm.}
\label{fig:MPC_block}
\end{figure}

Using the principle of superposition, and combining the velocity contributions from the uniform flow, the source and sink \eqref{eq:source_vel}, and the obstacle \eqref{eq:panel_velocity}, a collision-free velocity field can be generated. The resultant positional trajectory is obtained by integrating the velocity field and is then published to the drone's low-level position dynamic controllers. The drone's heading can be computed as follows for the low-level attitude controllers
\begin{equation}
    \omega_r = \tan^{-1}\left(\frac{v_y^r}{v_x^r}\right). 
\end{equation}
The integration of the VPM with the quadcopter control architecture is illustrated in Fig.~\ref{fig:MPC_block}.

\section{Feedback Controller Design with Collision Avoidance}

\subsection{System Modelling}
Consider a six-degree-of-freedom (6-DOF) quadcopter with unit vectors located at the center of mass, forming the rotation matrix $\mathbf{B} = [\Vec{\mathbf{b}}_x, \ \Vec{\mathbf{b}}_y, \ \Vec{\mathbf{b}}_z]\in SO(3)$. This matrix provides the transformation from the body-fixed reference frame to the north-east-down (NED) inertial reference frame $\{\Vec{i}_x, \ \Vec{i}_y, \ \Vec{i}_z\}$.
The translational dynamics are given by
\begin{align}\label{eq:position_dynamics}
    \dot{\bm{p}} &= \bm{v}, \\
    \dot{\bm{v}} &= g\bm{e}_3 - f\mathbf{B}\bm{e}_3,\label{eq:velocity_dynamics}
\end{align}
where $\bm{p} = [p_x, \ p_y, \ p_z]^T$ and $\bm{v}=[v_x, \ v_y, \ v_z]^T$ represent the position and velocity in the inertial frame, respectively. Here, $g$ is the acceleration due to gravity, $f$ is the mass-normalized collective thrust, $\bm{e}_3 = [0, \ 0, \ 1]^T \in \mathbb{R}^3$, and $-f\mathbf{B}\bm{e}_3\in \mathbb{R}^3$ denotes the total thrust in the inertial reference frame. 

The translational dynamics \eqref{eq:position_dynamics} and \eqref{eq:velocity_dynamics} can be described in a linear translation form as \cite{linear_translation}
\begin{align}\label{eq:linear_position_dynamics}
    \bm{\dot{p}} &= \bm{v}, \\
    \bm{\dot{v}} &= \bm{u},\label{eq:linear_velocity_dynamics}
\end{align}
where $\bm{u} = [a_{cmd}^x, \ a_{cmd}^y, \ a_{cmd}^z]^T$,  $a_{cmd}^x = fb_z^1$, $a_{cmd}^y = fb_z^2$, $a_{cmd}^z = g - fb_z^1$, with a superscript indicating an individual element of $\Vec{\mathbf{b}}_z$.

\subsection{High-Order Control Barrier Function}
In this section, some key results on HOCBFs that are necessary for our control design are briefly introduced (see \cite{HOCBF} for more details).

Consider a general affine control system of the form
\begin{equation}\label{eq:affine_system}
    \dot{\bm{x}} = h(\bm{x}) + q(\bm{x})\bm{u},
\end{equation}
where $\bm{x}\in \mathbb{R}^n$, $h:\mathbb{R}^n \rightarrow \mathbb{R}^n$, and $q:\mathbb{R}^n \rightarrow \mathbb{R}^n$ are globally Lipschitz, and $\bm{u} \in U \in \mathbb{R}^k$, where $U$ denotes a control set constraint. 
\begin{definition}
    A set $C \in \mathbb{R}^n$ is forward invariant for system \eqref{eq:affine_system} if, for some $\bm{u} \in U$, every solution starting at $\bm{x}(t_0)\in C$ satisfies $\bm{x}(t) \in C$, $\forall t \geq t_0$.
\end{definition}
Suppose that a safety constraint $b(\bm{x}) \geq 0$ is defined by an $m$-order differential function $b : \mathbb{R}^n \times [t_0, \infty) \rightarrow \mathbb{R}$ and let $\Gamma_0(\bm{x}) = b(\bm{x})$. We define a sequence of functions $\Gamma_i : \mathbb{R}^n \times [0,\infty)\rightarrow \mathbb{R}$, for $i \in \{1,\dots,m\}$: 
\begin{equation} \label{eq:seq_functions}
    \Gamma_i(\bm{x}) = \dot{\Gamma}_{i-1}(\bm{x}) + \chi_i(\Gamma_{i-1}(\bm{x})), \ i\in \{1,\dots,m\},
\end{equation}
where $\chi_i(\cdot)$ denote class $\mathcal{K}$ functions of their argument.

We further define a sequence of sets $C_i$, associated with \eqref{eq:seq_functions} in the form
\begin{equation}\label{eq:set}
    C_i = \{ \bm{x} \in \mathbb{R}  :  \Gamma_{i-1}(\bm{x}) \geq 0 \}, \ i\in \{1,\dots,m\}.
\end{equation}

\begin{definition} \emph{(\textbf{HOCBF} \cite{HOCBF})}\label{definition:HOCBF}
Let $C_i$, $i\in \{1,\dots,m\}$ be defined by \eqref{eq:set} and $\Gamma_i(\bm{x})$, be defined by \eqref{eq:seq_functions}. A function $b:\mathbb{R}^n \times [t_0,\infty)\rightarrow \mathbb{R}$ is a candidate HOCBF of relative degree $m$ for \eqref{eq:affine_system} if there exist class $\mathcal{K}$ functions $\chi_i$ such that
\begin{multline}\label{eq:HOCBF}\sup_{\bm{u} \in U} \left[\mathcal{L}^m_hb(\bm{x}) + \mathcal{L}_q\mathcal{L}^{m-1}_h b(\bm{x})\bm{u} + \frac{\partial^m b(\bm{x})}{\partial t^m}\right. \\ \left.+ O(b(\bm{x})) + \chi_m(\Gamma_{m-1}(\bm{x}))\right] \geq 0,
\end{multline}
for all $\bm{x} \in C_1\cap,\dots,\cap C_m \times [t_0,\infty)$. $\mathcal{L}_h$ and $\mathcal{L}_q$ denote the Lie derivatives along $h$ and $q$, respectively. $O(b(\bm{x})) = \sum_{i=1}^{m-1}\mathcal{L}_h^i(\chi_{m-i}\circ \Gamma_{m-i-1})(\bm{x})$. Further, $b(\bm{x})$ is such that $\mathcal{L}_q\mathcal{L}_h^{m-1}b(\bm{x}) \neq 0$ on the boundary of the set $C_1\cap,\dots,\cap C_m$.
\end{definition}

\begin{theorem}\emph{(\hspace{-0.0005cm}\cite{HOCBF})}\label{thm:HOCBF_thm}
Given a HOCBF $b(\bm{x})$ from Definition \ref{definition:HOCBF} with the associated sets $C_i$, $i \in \{1,\dots,m\}$ defined by \eqref{eq:set}, if $\bm{x}(t_0)\in C_1\cap,\dots,\cap C_m$, then any Lipschitz continuous controller $\bm{u}(t)$ that satisfies the constraint \eqref{eq:HOCBF}, $\forall t \geq t_0$ renders the set $C_1\cap,\dots,\cap C_m$ forward invariant for system \eqref{eq:affine_system}.
\end{theorem}

\subsection{HOCBF-Based Collision Avoidance Condition}

Consider the linear translational dynamics of the quadcopter, as described in \eqref{eq:linear_velocity_dynamics}. The goal is to synthesize a safety-critical controller that enforces dynamic collision avoidance constraints while simultaneously adhering to the softer navigation constraints generated by the VPM. By deriving the MPC solution based on the linear dynamics in \eqref{eq:linear_velocity_dynamics}, we assume they represent with sufficient accuracy the quadcopter's position evolution as governed by \eqref{eq:velocity_dynamics}. Control challenges such as dealing with disturbances, unknown dynamics, and modeling or measurement uncertainties can be addressed with robust MPC or event-triggered control frameworks. These methods can be integrated with the proposed VPM algorithm, and thus, a development of a multi-layer dynamic obstacle avoidance algorithm, VPM-MPC-HOCBF, is the primary focus of this work.

%robust MPC \cite{bao2023robust} or event-triggered control frameworks. These methods can be integrated with the proposed VPM algorithm, and thus, a development of a multi-layer dynamic obstacle avoidance algorithm, VPM-MPC-HOCBF, is the primary focus of this work.

%Unlike references \cite{xiao2023barriernet}[??,??,??,??] we are applying the HOCBF method to real-quadcopter systems, where assumptions on the influence of disturbances, time delay, measurement noise, modelling estimation inaccuracies in (??) are present. For these reasons, the HOCBF will not be applied as a hard constraint, as mathematical guarantees cannot always ensure safety on real systems.

\begin{figure}[bt]
\centerline{\includegraphics[width=0.95\columnwidth]{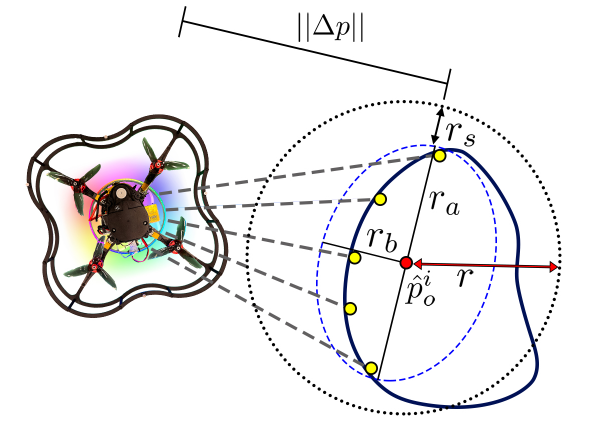}}
\caption{Geometric representation of a detected obstacle. The radial distance $r = r_a + r_s$, combined with the relative displacement $||\Delta \bm{p}||$, defines the safety set $b(\boldsymbol{\eta}) = ||\Delta \bm{p}||-r \geq 0$ for the HOCBF.}
\label{fig:ellipse}
\end{figure}

 For a navigating quadcopter with dynamics given by \eqref{eq:linear_position_dynamics} and \eqref{eq:linear_velocity_dynamics}, the relative position, velocity, and acceleration with respect to an obstacle moving along a smooth, continuous trajectory can be expressed as $\Delta \bm{p} = \bm{p} - \hat{\bm{p}}_{O}^i$, $\Delta \bm{v} = \bm{v} - \hat{\bm{v}}_{O}^i$, and $\Delta \bm{a} = \bm{u} - \hat{\bm{a}}_{O}^i$. The obstacle's dynamics are defined by $\hat{\bm{p}}_{O}^i = [x_O^i, \ y_O^i, \ z_O^i]$, with $\hat{\bm{v}}_{O}^i$ and $\hat{\bm{a}}_{O}^i$ defined analogously. These dynamics are to be estimated using real-time sensor data. The collision avoidance constraint between the quadcopter and the obstacle is expressed by the Euclidean distance $||\Delta \bm{p}|| \geq r$, where $r$ denotes the minimum safe distance that must be maintained from the obstacle, as illustrated in Fig. \ref{fig:ellipse}. Using the dynamics \eqref{eq:linear_position_dynamics} and  \eqref{eq:linear_velocity_dynamics}, a state vector can be formed as $\boldsymbol{\eta} = [\bm{{p}}, \ \bm{v}]$, and a HOCBF candidate is chosen as
\begin{equation}\label{eq:HOCBF_candidate}
    b(\boldsymbol{\eta}) = ||\Delta \bm{p}|| - r.
\end{equation}
%to ensure $||\Delta \bm{p}|| \geq r$ is maintained.

The relative degree of system \eqref{eq:linear_position_dynamics} and  \eqref{eq:linear_velocity_dynamics} is $m=2$, and $b(\boldsymbol{\eta})$ is used to define a series of functions $\Gamma_k$, $k=0,1,2$ of form \eqref{eq:seq_functions}. The class $\mathcal{K}$ functions \cite[Definition~1]{HOCBF} $\chi_1$ and $\chi_2$ are selected as linear functions, implying that:
\begin{equation}\label{eq:sequence_functions}
\begin{cases}
 \Gamma_0(\boldsymbol{\eta}) = b(\boldsymbol{\eta}), \\ \Gamma_1(\boldsymbol{\eta}) = 
     \dot{\Gamma}_0(\boldsymbol{\eta}) + \beta_1\Gamma_0(\boldsymbol{\eta}), \\ 
     \Gamma_2(\boldsymbol{\eta}) = 
     \dot{\Gamma}_1(\boldsymbol{\eta}) + \beta_2\Gamma_1(\boldsymbol{\eta}),
     \end{cases}
\end{equation}
where gains $\beta_1$, $\beta_2 \in \mathbb{R}^{+}$ are parameters of the HOCBF, and determine at what time the HOCBF constraint \eqref{eq:HOCBF} becomes active.

The sequence of sets associated with \eqref{eq:sequence_functions} are given as 
\begin{equation}
    \begin{cases}
        C_1 = \{\boldsymbol{\eta} \in \mathbb{R}  : \Gamma_0(\boldsymbol{\eta}) \geq 0 \}, \\
        C_2 = \{\boldsymbol{\eta} \in \mathbb{R} : \Gamma_1(\boldsymbol{\eta}) \geq 0 \}.
    \end{cases}
\end{equation}
\begin{comment}
consider a set $S \subset \mathbb{R}^n$ defined as the superlevel set of a continuously differntiable function $h:\mathbb{R}^n \rightarrow \mathbb{R}$ giving
\begin{equation}
    S = \{x \in \mathbb{R}^n \ | \ h(x) \geq 0\}.
\end{equation}

$S$ is referred to as the safe set, The function $h$ is defined as a control barrier function (CBF) if $\frac{\partial h}{\partial x} \neq 0$ for all $x \in \partial S$, and there exists an extended class $K_{\infty}$ function $\gamma$ such that $h$ satisfies
\begin{equation}
    \dot{h}(x,u) \geq -\gamma h(x).
\end{equation}

Will need to use  higher order control barrier function because we are dealig with a quadcopter system of relative degree 2 (a lower order is possible, but the higher order should suffice for this case I think). --> higher order is said to be conservative--> performance issues?? refer to the HOCBF book for the developments needed here 
\end{comment}
Considering the system dynamics \eqref{eq:linear_position_dynamics} with \eqref{eq:HOCBF_candidate} to compute the derivatives in the sequence of functions \eqref{eq:sequence_functions} gives the following
\begin{equation}\label{eq:gamma2}
\begin{cases}
    \Gamma_1(\boldsymbol{\eta}) &= \frac{\Delta \bm{p}^T \Delta \bm{v}}{||\Delta \bm{p}||} + \beta_1(||\Delta \bm{p}|| - r),\\
    \Gamma_2(\boldsymbol{\eta}) &= \Upsilon + \frac{\Delta \bm{p}^T}{||\Delta \bm{p}||}\Delta \bm{a}, \\
    \Upsilon &= \frac{||\Delta \bm{v}||^2}{||\Delta \bm{p}||} - \frac{(\Delta \bm{p}^T \Delta \bm{v})^2}{||\Delta \bm{p}||^3} + (\beta_1 + \beta_2)\frac{\Delta \bm{p}^T \Delta \bm{v}}{||\Delta \bm{p}||} \\ & \quad+ \beta_1\beta_2(||\Delta \bm{p}|| - r).
    \end{cases}
\end{equation}
\begin{comment}
    Use \cite[Definition 7]{HOBF} for reference
\end{comment}

Using \eqref{eq:HOCBF} and \eqref{eq:gamma2}, the HOCBF-based constraint is finally given as
\begin{equation} \label{eq:HOCBF_constraint}
    -\frac{\Delta \bm{p}^T}{||\Delta \bm{p}||}\begin{bmatrix}
        a_{cmd}^x - \hat{a}^i_{O_x} \\
        a_{cmd}^y - \hat{a}^i_{O_y} \\
        a_{cmd}^z - \hat{a}^i_{O_z}
    \end{bmatrix} \leq \Upsilon.
\end{equation}
Condition \eqref{eq:HOCBF_constraint} can be used as an inequality to constrain the quadcopter's control input, by which $b(\boldsymbol{\eta})>0$ is satisfied while tracking the trajectory generated by the VPM, in the presence of obstacles with complex dynamics.

%\subsection{Selection of the Gains $\beta_1$ and $\beta_2$}
\subsection[Selection of the Gains beta1 and beta2]{Selection of the Gains $\beta_1$ and $\beta_2$}

Selecting the HOCBF parameters $\beta_1$ and $\beta_2$ can be complex. Activating the HOCBF when the quadcopter is very close to an obstacle ($b(\boldsymbol{\eta}) \approx 0$) may require large, and potentially unfeasible control inputs. Conversely, it is also undesirable for the HOCBF constraint \eqref{eq:HOCBF_candidate} to activate too far from an obstacle, as early activation may cause the initial conditions of the HOCBF to have conflict with Theorem \ref{thm:HOCBF_thm}.

An optimal parameter selection is beyond the scope of this work. Readers can refer to \cite{HOCBF} and \cite{xiao2023barriernet} for detailed methods on optimal parameter selection. In this work, parameters were tuned manually, with a large range of values giving satisfactory results in both simulation and hardware tests.

%Weights $\beta_1$, $\beta_2 \in \mathbb{R}^{+}$ limit the demand of the HOCBF constraint (??) for control input $\textbf{u}$ after it becomes active.... 

%Although it is recommended to set the HOCBF as hard constraints for safety \cite{xiao2023barriernet},\cite{manip_STO}; slack variables are included in the ACADOS solver to relax the constraints (??). This is common practice for real system application, because unlike idealized mathematical simulations, you cannot guarantee real robotic systems will maintain the defined safety sets (??) given the control input $\textbf{u}$ from the OCP (??). This is due to limited control authority, hardware limitations, unmodelled system dynamics and disturbances (assimung the linear dynamic model(??) represents the nonlinear translational dynamics (??)). This is illustrated in the PX4-power Gazebo simulations along with hardware testing in Sections ?? and ?? respectively. Softening the constraints avoids running into infeasible solver errors, particularly SQP-RTI errors in the ACADOS solver, which would put the Clover drone into failsafe mode.

\subsection{Model Predictive Control}

Building on the reference field $\bm{v}_r = [v_x^r, \ v_y^r, \ v_z^r]^T$ and the corresponding reference trajectory $\bm{p}_r = [p_x^r, \ p_y^r, \ p_z^r]^T$ obtained from the VPM, let $\bm{e} = \bm{p} - \bm{p}_r$ denote the trajectory tracking error between the Clover quadcopter and the reference. The general form of the finite-horizon MPC formulation for quadcopter streamline tracking, subject to the HOCBF constraints in \eqref{eq:HOCBF_constraint} at time $t_k$, is given as follows 
%\begin{align}\label{eq:MPC_continuous}
%    \min_{\bm{u}} \int_{t_k}^{t_k+T} \hspace{-1em} &||\bm{e}(s)||^2_{\bm{W}} + ||\bm{u}(s)||^2_{\bm{R}} \,ds + ||\bm{e}(t_k + T)||^2_{\bm{P}} \\
%   \underset{\text{\normalsize to}}{\text{subject}} \ \ &\dot{\boldsymbol{\eta}}(s) = f(\boldsymbol{\eta}(s),\bm{u}(s)), \ s = [t_k, t_k+T) \\
%   &\boldsymbol{\eta}(0) = \boldsymbol{\eta}_{init}, \ 
%    \bm{u}(s) \in [\bm{u}_{min} \ \bm{u}_{max}] \label{eq:mpc_in_con_1} \\ 
%    &-\frac{\Delta \bm{p}^T(s)}{||\Delta \bm{p}(s)||}\Delta \bm{a}(s) \leq \boldsymbol{\Upsilon}(s) + \boldsymbol{\sigma}(s)  \label{eg:mpc_in_con_2}
%\end{align}
\begin{align}\label{eq:MPC_continuous}
    \min_{\bm{u}} \int_{t_k}^{t_k+T} \hspace{-1em} &||\bm{e}(s)||^2_{\bm{W}} + ||\bm{u}(s)||^2_{\bm{G}} \,ds + ||\bm{e}(t_k + T)||^2_{\bm{P}} \\
    \underset{\text{\normalsize to}}{\text{subject}} \ \ 
    &\dot{\bm{p}}(s) = \bm{v}(s), \ \dot{\bm{v}}(s) = \bm{u}(s), \ s \in [t_k, t_k + T) \label{eq:mpc_dynamics} \\
    &\boldsymbol{\eta}(0) = \boldsymbol{\eta}_{\text{init}}, \quad 
    \bm{u}(s) \in [\bm{u}_{\min}, \bm{u}_{\max}] \label{eq:mpc_in_con_1} \\ 
    &-\frac{\Delta \bm{p}^T(s)}{||\Delta \bm{p}(s)||}\Delta \bm{a}(s) \leq \boldsymbol{\Upsilon}(s) + \boldsymbol{\sigma}(s)  \label{eg:mpc_in_con_2}
\end{align}
where $\bm{W}, \bm{G}, \bm{P} \in \mathbb{R}^{3x3}$ are positive definite weight matrices for the states and inputs, respectively. Variable $\boldsymbol{\sigma}$ is a slack vector which relaxes the HOCBF constraint to guarantee the feasibility of solutions.

The problem is discretized into $N$ steps over a time horizon $T$, with a step size of $\delta_t = T/N$, defining the prediction horizon. The dynamics in \eqref{eq:linear_position_dynamics} and \eqref{eq:linear_velocity_dynamics} are evaluated using an explicit fourth-order Runge-Kutta method to propagate the state forward. At each timestep $k$, given the current state $\boldsymbol{\eta}_k$ and control input $\bm{u}_k$, the next step $\boldsymbol{\eta}_{k+1}$ is computed by numerically integrating the system over the interval $[t_k, \ t_k+\delta_t]$.

%The problem is discretized into $N$ steps over a time horizon $T$, with a step size of $\delta_t = T/N$, defining the prediction horizon. The dynamics in \eqref{eq:affine_system} are evaluated using an explicit 4th-order Runge-Kutta method, denoted as ${f}_{RK4}$, which integrates $\dot{\boldsymbol{\eta}}$ given an initial state $\boldsymbol{\eta}_k$, control input $\bm{u}_k$, and integration step $\delta_t$, resulting in $\boldsymbol{\eta}_{k+1} = {f}_{RK4}(\boldsymbol{\eta}_k, \bm{u}_k, \delta_t)$.

%The problem is descritizd into $N$ steps over a time horizon $T$ of size $\delta t = T/N$, giving the prediction horizon. The dynamics in \eqref{eq:affine_system} are evaluated with an explicit Runge-Kutta method of 4th order $\textbf{f}_{RK4}$ to integrate $\dot{\textbf{x}}$ given an initial state $\textbf{x}_k$, control input $\textbf{u}_k$, and integration step $\delta t$ with $\textbf{x}_{k+1} = \textbf{f}_{RK4}(\textbf{x}_k, \textbf{u}_k,\delta t)$.
\begin{comment}
The problem \eqref{eq:MPC_continuous} is solved at each sampling time $t_k$:
\begin{align}
    \min_{\textbf{u}} J = ||\textbf{e}_N||^2_{\textbf{Q}_N} + \sum_{i=0}^{N-1}(||\textbf{e}_i||^2_{\textbf{Q}} + ||\textbf{u}_i||^2_{\textbf{R}}) \\
    subject \  to \ \ \textbf{x}_{k+1} = \textbf{f}_{RK4}(\textbf{x}_k, \textbf{u}_k,dt), \ \ \textbf{x}_0 = \textbf{x}_0 \\
    \mathbf{u} \in [\mathbf{u}_{min} \ \mathbf{u}_{max}]
\end{align}
where $Q, R, Q_N \in \mathbb{R}^{3x3}$ are positive definite weight matrices for the states and inputs respectively.
\end{comment}

To solve the nonlinear optimization problem at each timestep $t_k$, the optimal control problem (OCP) in \eqref{eq:MPC_continuous}-\eqref{eg:mpc_in_con_2} is discretized using a direct multiple shooting method, resulting in a finite-dimensional nonlinear problem. This problem is then solved in a receding horizon framework using sequential quadratic programming (SQP) within a real-time iteration (RTI) scheme. All implementations are carried out using the ACADOS \cite{acados} software.

\section{Obstacle State Estimation}
\subsection{Obstacle Trajectory Model}\label{sec:obstacle_model}

This section presents the general model used to propagate dynamic obstacle trajectories forward within the prediction horizon $T$ for the optimal control problem \eqref{eq:MPC_continuous}-\eqref{eg:mpc_in_con_2}. As discussed in \cite{NMPC_obsavoidance_mulTtraj}, obstacles may follow various trajectory types, including linear and projectile motion. More complex paths, such as elliptic, parabolic, hyperbolic, helical, and spiral trajectories, can also arise. However, predicting and modeling obstacles with random or intricate motion patterns remains challenging, as their objectives are often unknown.

To better capture and predict obstacle trajectories, particularly when they are within LiDAR range and moving along high-acceleration paths (e.g., other robots), this work employs linear acceleration trajectory models.

The linear trajectory model for the \emph{i}th obstacle is represented by
\begin{equation}\label{eq:linear_acceleration}
\begin{cases}
\begin{aligned}
    \hat{\bm{p}}^i_{O}(k) &= \hat{\bm{p}}^i_{O}(k-1) + \hat{\bm{v}}^i_O(k-1) t + \frac{1}{2}\hat{\bm{a}}^i_O(k-1) t^2, \\
    \hat{\bm{v}}^i_O(k) &= \hat{\bm{v}}^i_{O}(k-1) + \hat{\bm{a}}^i_O(k-1) t, \\
    \hat{\bm{a}}^i_O(k) &= \hat{\bm{a}}^i_O(k-1),
    \end{aligned}
    \end{cases}
\end{equation}
where $(\bm{p}^i_{O},  \bm{v}^i_{O}, \bm{a}^i_O)$ denote the position, velocity, and acceleration of the detected obstacle. Since these quantities are not directly observable, they are estimated as $(\hat{\bm{p}}^i_{O}, \hat{\bm{v}}^i_{O}, \hat{\bm{a}}^i_O)$ using global LiDAR measurements $\bm{Z}_k = [\mathbf{x}_{\mathcal{G}}, \mathbf{y}_\mathcal{G}]$ of surface $\mathcal{G}$ and ellipse-based shape estimation, as illustrated in Fig. \ref{fig:ellipse}. 

%$\mathbf{x} = [x_{O}, y_{O},\dot{x}_{O}, \dot{y}_{O}, \ddot{x}_{O}, \ddot{y}_{O}, a,b,\theta] \in \mathbb{R}^9$,

\subsection{AKF State Estimation}\label{sec:AKF}

The ellipse shape parametrization of detected obstacles is performed using the MBE algorithm presented in \cite{welzl2005smallest}. To predict obstacle states in the presence of sensor noise, an AKF is designed which integrates MBE-based observations. The state vector of the dynamic model is $\mathbf{x} = [\bm{p}^i_{O},\bm{v}^i_{O}, {\bm{a}}^i_O, \boldsymbol{\Sigma}]^T \in \mathbb{R}^9$, where $\boldsymbol{\Sigma}=[r_a,r_b,\theta]^T$ represents the ellipse shape parameters, with $r_a$ and $r_b$ denoting the major and minor axes (see Fig. \ref{fig:ellipse}), and $\theta$ representing the ellipse's orientation angle. The AKF estimation process, incorporating system dynamics \eqref{eq:linear_acceleration} and MBE observations, is given by
\begin{equation}
    \begin{cases}
    \begin{aligned}
    \hat{\mathbf{x}}^{-}(k) &= \bm{A}(k)\hat{\mathbf{x}}(k-1),\\
    \bm{P}^{-}(k) &= \bm{A}(k)\bm{P}(k-1)\bm{A}^{T}(k) + \bm{Q}(k),\\
    \Tilde{\bm{\epsilon}}(k) &= \bm{z}(k) - \bm{H}(k)\hat{\mathbf{x}}^{-}(k),
    \end{aligned}
    \end{cases}
\end{equation}
where $\Tilde{\bm{\epsilon}}(k)$ is the innovation, $\bm{z}(k) = [\bm{p}^i_O(k), \boldsymbol{\Sigma}(k)]^T \in \mathbb{R}^5$ is the measurement, $\hat{\mathbf{x}}^{-}(k)$ and $\bm{P}^{-}(k)$ are the priori state estimate and covariance matrix, while $\hat{\mathbf{x}}(k-1)$ and $\bm{P}(k-1)$ are the posteriori state estimate and covariance matrix, respectively. $\bm{Q}(k)$ represents the process noise covariance matrix. The state transition matrix $\bm{A}(k)$ and measurement matrix $\bm{H}(k)$ are given as
\begin{align} \label{eq:KF_matrices}
    \bm{A}(k) &= \begin{bmatrix}
        1 & 0 & t & 0& \frac{t^2}{2} & 0\\
         &1& 0 & t & 0 & \frac{t^2}{2} \\
        & & 1 & 0 & t & 0 & \bm{O}_{6x3}\\
        \vdots& & & 1 & 0 & t\\
        & & & & 1 & 0\\
        0& &\cdots & & & 1\\
        & & \bm{O}_{3x6} & & & & \bm{I}_{3x3}
    \end{bmatrix},\\
    \bm{H}(k) &= \begin{bmatrix}
        \bm{I}_{2x2} & \bm{O}_{2x7}\\
        \bm{O}_{3x6} & \bm{I}_{3x3}
    \end{bmatrix}.
\end{align}
Using $\bm{P}^{-}(k)$, the Kalman gain is computed as follows
\begin{equation}
    \bm{K}(k) =  \bm{P}^{-}(k)\bm{H}^{T}(k)[\bm{H}(k)\bm{P}^{-}(k)\bm{H}^{T}(k) + \bm{R}(k)],
\end{equation}
where $\bm{R}(k)$ is the measurement noise covariance matrix. The state estimates are then updated with
\begin{equation}
    \begin{cases}
    \begin{aligned}
        \hat{\mathbf{x}}(k) &= \bm{A}(k)\hat{\mathbf{x}}^{-}(k) + \bm{K}(k)\Tilde{\bm{\epsilon}}(k), \\
        \bm{P}(k) &= [\bm{I} - \bm{K}(k)\bm{H}(k)]\bm{P}^{-}(k),\\
        \Tilde{\bm{y}}(k) &= \bm{z}(k) - \bm{H}(k)\hat{\mathbf{x}}(k),
        \end{aligned}
    \end{cases}
\end{equation}
where $\hat{\mathbf{x}}(k)$ and $\bm{P}(k)$ are the posteriori state estimate and covariance matrix, respectively, while $\Tilde{\bm{y}}(k)$ is the measurement residual.

\subsection{Adaptive State Estimation}

The change in position over time is a result from both the motion of the dynamic obstacle and adjustments in the MBE fitting. Frequent adjustments from the MBE algorithm can cause rapid variations in the estimated ellipse center. 
To mitigate these variations, an adaptive covariance matrix \cite{AKF} is implemented, taking the following form
\begin{align}
    \bm{R}(k) &= \alpha \bm{R}(k-1) + (1-\alpha)(\Tilde{\bm{y}}(k) \Tilde{\bm{y}}(k)^T \nonumber \\ &\quad+ \bm{H}(k) \bm{P}^{-}(k) \bm{H}(k)^T),
\end{align}
and an adaptive estimation for $\bm{Q}(k)$ is given as
\begin{equation}
    \bm{Q}(k) = \alpha \bm{Q}(k-1) + (1-\alpha)(\bm{K}(k) \Tilde{\bm{\epsilon}}(k) \Tilde{\bm{\epsilon}}(k)^T \bm{K}(k)^T),
\end{equation}
where $\alpha\in(0,1)$ is a forgetting factor. A larger $\alpha$ places greater weight on the previous estimates, reducing fluctuations in $\bm{R}(k)$ and $\bm{Q}(k)$, but introducing slower adaptation to changes. To balance stability and responsiveness, an adaptive forgetting factor is designed as follows
\begin{align}
    \alpha &= \alpha_{min} + (\alpha_{max}-\alpha_{min})(1 - e^{-\varrho\Tilde{\boldsymbol{\Sigma}}}),\\
    \Tilde{\boldsymbol{\Sigma}} &= \boldsymbol{\Sigma} - \Bar{\boldsymbol{\Sigma}},
\end{align}
where $\varrho$ is 
a tunable gain, $\Bar{\boldsymbol{\Sigma}}$ represents the exponential moving average of $\boldsymbol{\Sigma}$, and $\Tilde{\boldsymbol{\Sigma}}$ quantifies its deviation from the average. When $\boldsymbol{\Sigma}$ changes rapidly, $\Tilde{\boldsymbol{\Sigma}}$ increases, leading to a larger covariance matrix $\bm{R}(k)$. This places greater weight on previous estimates, reducing the sensitivity to fluctuations in $\boldsymbol{\Sigma}$.

\subsection{Analysis of Uncertainty}\label{sec:uncertainty}

At each iteration of the MPC control problem in \eqref{eq:MPC_continuous}, obstacle dynamics must be predicted over the prediction horizon. To accomplish this, an open-loop forward model initialized with the latest estimate from the AKF in Section \ref{sec:AKF} propagates the obstacle dynamics across the horizon, updating each step with
\begin{equation}
    \begin{cases}
        \begin{aligned}
            \hat{\mathbf{x}}_{O}(k) &= \bm{A}_{O}(k)\hat{\mathbf{x}}_{O}(k-1), \\
            \bm{P}_{O}(k) &= \bm{A}_{O}(k)\bm{P}_{O}(k-1)\bm{A}_{O}^{T}(k) + \bm{Q}_{O}(k),
        \end{aligned}
    \end{cases}
\end{equation}
where $\mathbf{x}_O = [\bm{p}^i_{O},\bm{v}^i_{O},{\bm{a}}^i_O]^T \in \mathbb{R}^6$. The covariance matrix $\bm{P}_{O}(k)$ captures the uncertainty associated with each prediction. 

Using the position covariance $\bm{P}_{\bm{p}^i_O}$, an uncertainty region can be generated based on the Mahalanobis distance or the standard deviation of position. The maximum uncertainty is given by
\begin{equation}
    \sigma_p = \sqrt{\lambda_{\max}\left(\bm{P}_{\bm{p}^i_O}\right)},
\end{equation}
where $\lambda_{\max}$ is the maximum eigenvalue, corresponding to the largest uncertainty in the position for a given iteration. To mitigate the risk associated with uncertainty, the safety distance $r_s$, defined in Fig. \ref{fig:ellipse}, is adjusted along the prediction horizon as follows
\begin{equation}
    r_{s_i} = r_{s} + \Lambda_0 \sigma_p, \  i\in \{0,...,N\},
\end{equation}
where $\Lambda_0$ is a scaling factor applied to the Mahalanobis distance. The scaling factor can be set based on Gaussian distribution uncertainty or slightly increased to account for additional uncertainties.

\section{Simulation Studies}\label{sec:simulations}

To demonstrate the effectiveness of the reactive obstacle avoidance algorithm, we conducted simulations using PX4's software-in-the-loop (SITL) with the Clover-Gazebo simulator, which replicates real Clover hardware and integrates ROS. The Gazebo models incorporate system geometry, inertial properties, and more realistic physics compared to simulations based on simplified differential equations. The SITL Clover package includes sensors such as a laser rangefinder and a camera, with Gaussian noise models simulating sensor inaccuracies. A 360$^{\circ}$ Laser Distance Sensor (LDS-01), providing 360 samples per scan (1$^\circ$ resolution) with a range of $3.5\,\text{m}$, is integrated with the Clover in the Gazebo simulation environment.

%To demonstrate the effectiveness of the reactive obstacle avoidance algorithm, we conducted simulations using PX4's software-in-the-loop (SITL) feature. This involved a PX4-powered Clover-Gazebo simulator environment the replicates real Clover hardware, integrated with ROS. The Gazebo models incorporate system geometry and inertial properties, enabling more realistic physics compared to simplified differential equation-based simulations. The SITL Clover package includes sensors used on a real Clover drone, such as a laser rangefinder, optical flow, and ArUco marker-based computer vision. Gaussian noise models replicate real sensor measurements. A 360$^{\circ}$ Laser Distance Sensor (LDS-01), providing 360 samples per scan (1$^\circ$ resolution) with a range of $3.5\,m$, was integrated with the Clover in the Gazebo simulation environment.

The VPM outlined in Section \ref{sec:VPM} is implemented in Python as a robot operating system (ROS) node, generating collision-free trajectory setpoints to the onboard PX4 control architecture via MAVROS. The MPC-HOCBF solver, developed in ACADOS, tracks these setpoints while incorporating LiDAR-based AKF feedback. MPC commands are sent to PX4 as acceleration setpoints via MAVROS.

To isolate the contributions of the VPM and MPC-HOCBF-AKF, simulations varied the VPM update frequency. In the VPM-MPC-HOCBF-AKF configuration, the VPM continuously updated upon obstacle detection. In contrast, in the VPM*-MPC-HOCBF-AKF configuration, the initial obstacle-free VPM-generated velocity field was used for tracking, while avoidance was handled solely by MPC-HOCBF-AKF. This methodology was applied to both static and dynamic test cases. All of the design parameters for the proposed navigation and control methods are given in Table \ref{table:ctrlparam}.

%The VPM outlined in Section \ref{sec:VPM} was developed in Python and integrated into a ROS node, which communicates robust collision-free trajectory setpoints to the onboard PX4 control architecture via MAVROS. Similarly, the MPC-HOCBF solver was developed in ACADOS using the Python interface, communicating with the VPM for reference tracking setpoints. The AKF, utilizing LiDAR feedback, was combined to provide feedback for the MPC-HOCBF problem. MPC commands were sent to the PX4 control architecture as acceleration setpoints via MAVROS.

%To evaluate obstacle avoidance performance, the following metrics are used:
The obstacle avoidance performance is evaluated using:
\begin{itemize}
    \item Speed variance, measured when obstacles are within LiDAR range.
    \item Control effort, computed as $\int_{t_0}^{t_f}  \bm{u}^2 dt$, assessing control input efficiency.
    \item Minimum distance to obstacles, with static cases considering both the closest approach and average minimum distance, while dynamic cases use only the closest approach.
\end{itemize}
The performance metric results are presented in Table \ref{table:mpc_dynamic}. 

%\begin{itemize}
%    \item VPM-MPC-HOCBF-AKF: VPM continuously updates upon obstacle detection.
%    \item VPM*-MPC-HOCBF-AKF:The initial VPM-generated velocity field is used for tracking, with avoidance handled solely by MPC-HOCBF-AKF.
%\end{itemize}
%This methodology was applied to both static and dynamic test cases.

%The results are shown in Table \ref{table:mpc_dynamic}. To separately assess the benefits of the VPM and MPC-HOCBF-AKF components, various simulation scenarios were conducted. In some simulations, the VPM is continuously updated upon obstacle detection (denoted as VPM-MPC-HOCBF-AKF), while in others, a velocity field initialized by the VPM is used for tracking, and only the MPC-HOCBF-AKF algorithm handles avoidance (denoted as VPM*-MPC-HOCBF-AKF). This approach was applied to both static and dynamic test cases.

\begin{table}[bp]
\caption{Control and navigation design parameters for each algorithm used in the simulations and experiments.}
\centering
\begin{tabular}{c|c}
\hline
\multicolumn{2}{c}{Gazebo Simulations}\\
 \hline
 Algorithm & Parameters \\
 \hline \hline
 VPM-A & $\mu = 0.3$, $\kappa = 0^{\circ}$, $\ell_{kutta} = 0.8\,\text{m}$  \\
 VPM-B & $\xi = \pm\{0.3,0.5\}$   \\
 VPM-MPC-HOCBF & 2D: $\Lambda_0 = 2$, $\beta_1 = 4$, $\beta_2 = 4$\\
   & 3D: $\Lambda_0 = 2$, $\beta_1 = 1$, $\beta_2 = 2$  \\
 AKF & $\alpha_{\min}=0.7$, $\alpha_{\max}=1$, $\varrho=1.5$\\
 \hline
 \multicolumn{2}{c}{Hardware Experiments}\\
 \hline
 Algorithm & Parameters\\
 \hline\hline
 VPM-A & $\mu = 0.3$, $\kappa = 10^{\circ}$, $\ell_{kutta} = 0.15\,\text{m}$\\
 VPM-B & $\xi = \pm\{0.3,0.4,0.5\}$ \\
 VPM-MPC-HOCBF & $\Lambda_0 = 2$, $\beta_1 = 1$, $\beta_2 = 2$\\
 AKF & $\alpha_{\min}=0.7$, $\alpha_{\max}=1$, $\varrho=1.2$\\
 \hline
\end{tabular}
\label{table:ctrlparam}
\end{table}

% \begin{table}[h]
%     \caption{Simulation and experimental HOCBF and observer parameters.}
%     \centering 
%     \begin{tabular}{c c c c}
%         \hline
%         Eq. \eqref{eq:adptlaw} - Sim & $\beta_1 = 13$ & $\beta_2 = 9$ & - \\ 
%         & $\mu_{\alpha_t} = 0.001$ & $\mu_{\alpha_r} = 0.005$ & $\eta_\alpha = 0.1$  \\ \hline
%         Eq. \eqref{eq:adptlaw} - Exp & $\beta_1 = 4$ & $\beta_2 = 3$ & - \\ 
%         \hline
%         HOSMO & \multicolumn{3}{l}{$\lambda_1=\lambda_2=\lambda_3 = 13$} \\ \hline
%     \end{tabular}
% \end{table}

A video of the simulations and experiments can be viewed at \emph{\url{https://youtu.be/NqSFB2RTiEU}}.

\subsection{Static Obstacles}
To evaluate the VPM algorithm for static obstacle avoidance, a complex obstacle array is created, as shown in Fig. \ref{fig:VPM_static}. The APF algorithm \cite{khatib1986real} serves as the baseline, and the VPM-MPC-HOCBF-AKF is evaluated in comparison under the same obstacle configuration.

\subsubsection{Online Vortex Panel Method}\label{sec:VPM_gaz_static}

\begin{figure}[tb]
\centerline{\includegraphics[width=0.95\columnwidth]{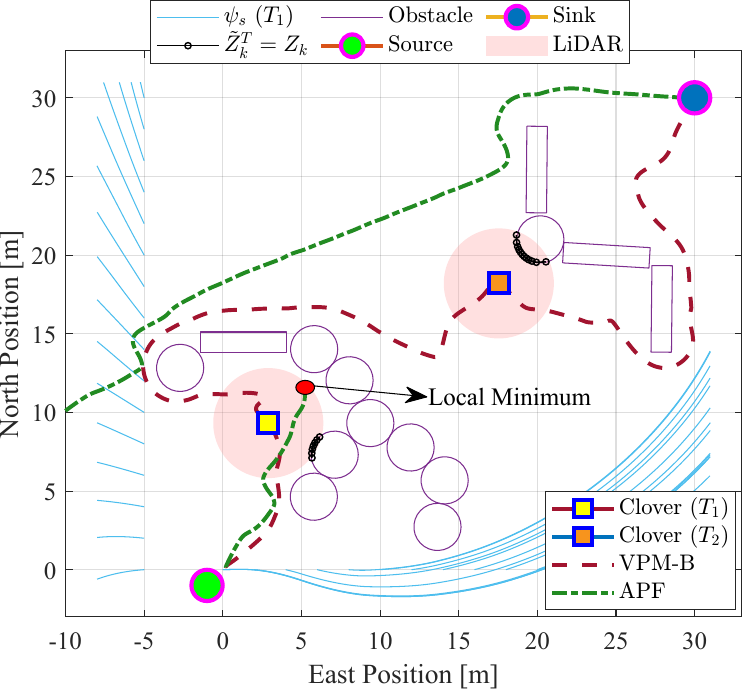}}
\caption{Clover Gazebo simulation showing global trajectory profiles generated by the VPM-B algorithm and the APF algorithm. The resultant velocity field, with a parameter $\xi= \pm 0.3$, guides the Clover drone to its destination while avoiding two complex static obstacles. Two timestamps, $T_1 = 18.85\,\text{s}$ and $T_2=63.46\,\text{s}$, along the trajectory are displayed.}
\label{fig:VPM_static}
\end{figure}

In Fig. \ref{fig:VPM_static}, the Clover's trajectory, navigating from the source to the sink using the VPM-B algorithm with $\xi = \pm 0.3$, is shown. Initially, the velocity field is defined using only the source and sink in an obstacle-free environment. A high-level ROS node for VPM-B calculations remains on standby until LiDAR detects an obstacle. At time $T_1$, LiDAR identifies a surface, updating the streamlines $\psi_s(T_1)$ based on scan data $\Tilde{\bm{Z}}^T_k = \bm{Z}_k$, which reshapes the velocity field to guide the Clover around the concave obstacle. In contrast, the APF method directs the Clover into the concave region, resulting in an auto-landing due to a local minimum.

%In Fig. \ref{fig:VPM_static}, the Clover's trajectory, navigating from the source to the sink using the VPM-B algorithm with $\xi = \pm 0.3$, is shown. The navigation begins by initializing a velocity field using only the source and sink components in an obstacle-free environment. A high-level ROS node, responsible for VPM-B calculations, remains on standby until an obstacle is detected by the LiDAR sensor. Upon reaching position $O_1$, the LiDAR detects the surface. The scan reading, $\Tilde{\bm{Z}}^T_k = \bm{Z}_k$ updates the streamlines $\psi_s(T_1)$, providing a velocity field that guides the Clover out and around the concave obstacle. In contrast, the APF method directs the Clover into the concave region, forcing it to auto-land upon reaching a local minimum.

To reduce computational load, the VPM-B pauses when no obstacles are detected, reactivating only upon new LiDAR detections. At $T_2$, corresponding to the velocity field in Fig. \ref{fig:VPM_static}, the second obstacle is detected, again updating $\Tilde{\bm{Z}}^T_k = \bm{Z}_k$. With $\xi = 0.3$, the sink dominates the flow field, guiding fluid particles and the Clover around obstacles toward the goal.

\begin{table}[bp]
\caption{Metrics used to compare the performance of the proposed algorithm with baseline methods. For static cases, the minimum and mean minimum distances are reported as $[\text{min}, \text{mean}]$. For dynamic cases, only the minimum distance is reported.}
\centering
\begin{tabular}{ c|>{\centering}p{1.1cm}|>{\centering}p{1.1cm}|>{\centering\arraybackslash}p{1.5cm}}
\hline
\multicolumn{4}{c}{Gazebo Simulations}\\
 \hline
 Algorithm & Speed Variance $[\text{m/s}]^2$ & Control Effort $\left[ {\text{m}^2}/{\text{s}^3} \right]$ & Minimum Distance $[\text{m}]$ \\
 \hline 
 \multicolumn{4}{c}{Static}\\
 \hline
 VPM-B ($\xi = \pm 0.3$) & 0.0326 & \textemdash & [0.58, 2.49] \\
 VPM-MPC-HOCBF-AKF & 0.0357  & 20.23  & [0.68, 2.73]   \\
 \hline
 \multicolumn{4}{c}{Complex Dynamic}\\
 \hline
 VPM*-MPC-HOCBF-AKF & 0.1071  & 522  & 0.77   \\
 VPM*-MPC-AKF & 0.1174  & 830  & 0.72   \\
 VPM*-MPC & 0.2694  & 711  & 0.34  \\
 \hline
 \multicolumn{4}{c}{3D Dynamic}\\
 \hline
 VPM*-MPC-HOCBF-AKF & 0.2156  & 66.7  & 0.85   \\
 VPM*-MPC-AKF & 0.3176  & 92.5  & 0.78   \\
 VPM*-MPC & 0.3717  & 85.5  &  0.23  \\
 \hline
\end{tabular}
\label{table:mpc_dynamic}
\end{table}

\begin{figure}[tb]
\centerline{\includegraphics[width=0.95\columnwidth]{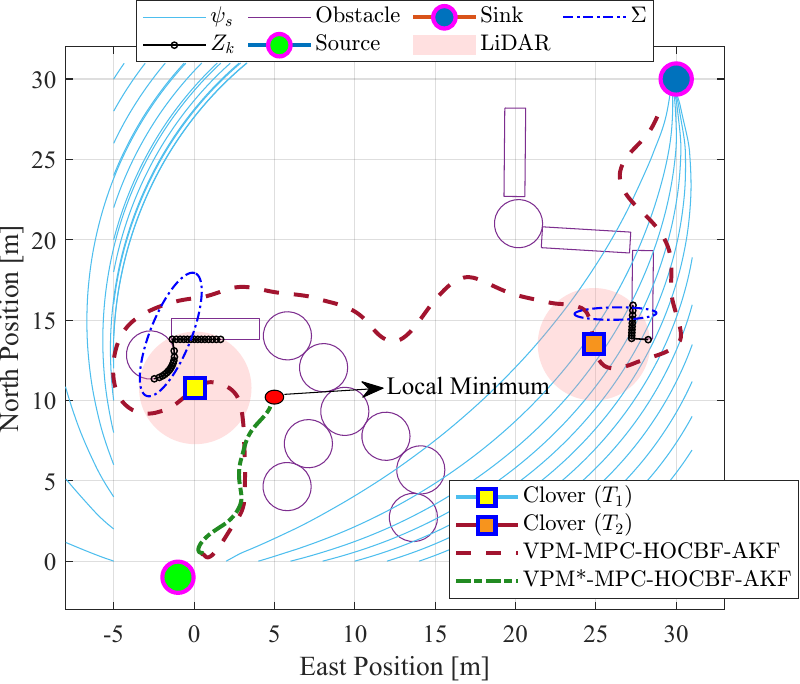}}
\caption{Clover Gazebo simulations showing global trajectory profiles and obstacle avoidance generated by the VPM-MPC-HOCBF-AKF algorithm while avoiding two complex static obstacles. Two timestamps, $T_1 = 19.97\,\text{s}$ and $T_2=69.86\,\text{s}$, along the trajectory are displayed.}
\label{fig:MPC_static}
\end{figure}

\subsubsection{VPM-MPC-HOCBF-AKF}\label{sec:MPC_static}
%Influence of beta1,beta2 values? From hardware it seemed reducing them allowed the clover to react later, not too early. Refer to the other reference similar to yours and see if they mention it... maybe include a section on the selction, and relateit bac to th eoriginal HOCBF paper

%For this simulation, the velocity field is initialized using the source, sink, and an unknown environment, providing trajectory setpoints used as the tracking components for the MPC optimal problem defined in \eqref{eq:MPC_continuous}. An optimal acceleration control input is provided to the Clover's PX4 control system, aimed at tracking the generated trajectory and managing the control input constraints \eqref{eq:mpc_in_con_1} and  \eqref{eg:mpc_in_con_2}. 
In Fig. \ref{fig:MPC_static}, the trajectory of the Clover, navigating from the source to the sink, is shown. At time $T_1$, the LiDAR readings $\bm{Z}_k$ are used to fit an MBE, serving as measurement feedback for the AKF. The resulting output ellipse $\boldsymbol{\Sigma}$ from the AKF is shown. The AKF provides obstacle state feedback for the HOCBF constraint \eqref{eg:mpc_in_con_2} within the MPC problem, ensuring collision avoidance. Without a quasi-updating VPM algorithm, the VPM*-MPC-HOCBF-AKF lacks the logic to escape the concave regions, leading to either entrapment or collision, depending on the controller weight selection.

%In Fig. \ref{fig:MPC_static}, at position $O_1$, the LiDAR readings $\bm{Z}_k$ are used to fit an MBE, which serves as an observation and measurement feedback for the AKF. The current output ellipse $\boldsymbol{\Sigma}$ from the AKF is shown. The AKF provides obstacle pose parameters for the HOCBF constraint \eqref{eg:mpc_in_con_2} within the MPC problem, ensuring collision avoidance. Without a quasi-updating VPM algorithm, the MPC-HOCBF-AKF lacks the logic to escape the concave region and would either get stuck within the region or collide with the obstacle, depending on the selection of optimal controller weights.

Having a longer prediction horizon ($T=5\,\text{s}$ compared to $T=1\,\text{s}$) allows the MPC to steer the Clover system to avoid obstacles earlier, thereby improving both reaction time and flight stability. Compared to the VPM in Section \ref{sec:VPM_gaz_static}, the MPC-HOCBF-AKF introduces an additional layer of control and safety by actively managing the distance from obstacles. This improvement is reflected by a $9.6\,\%$ increase in the mean minimum distance, as shown in Table \ref{table:mpc_dynamic}.

\subsection{Dynamic Obstacles}
To evaluate the effectiveness of the proposed algorithm for dynamic obstacle avoidance, various dynamic cases are considered. Initially, simple dynamic obstacles are used to assess the robustness of the VPM, followed by improvements achieved with integrating the MPC-HOCBF-AKF as an additional control layer. Baseline methods for comparison include:

%To test and demonstrate the benefits of the developed algorithm for dynamics obstacle avoidance, various dynamic cases are considered. First, simple dynamic obstalces are use to study the robustness of the VPM, followed by improvements shown by integrating the MPC-HOCBF-AKF as an additional layer of control. Baseline methods from comparison are
\begin{itemize}
    %\item MPC: MPC tracking controller considering safety constraints with Euclidean norm distance \cite{NMPC_obsavoidance_mulTtraj,MPC_GP}
    \item VPM*-MPC: a standard MPC tracking controller with Euclidean norm safety constraints \cite{NMPC_obsavoidance_mulTtraj,MPC_GP};
    %\item MPC-AKF: MPC with a Euclidean norm distance, where the obstacle state evolution within the prediction horizon is handled using the proposed AKF, which considers obstacle acceleration as opposed to \cite{MPC_GP,MPC_kalman,MPC_EKF_gauss}.
    \item VPM*-MPC-AKF: an MPC with Euclidean norm constraints and obstacle state evolution \cite{MPC_GP,MPC_kalman,MPC_EKF_gauss}, improved by accounting for obstacle acceleration using the proposed AKF.
\end{itemize}
In dynamic simulations with the MPC layer, the VPM is initialized for trajectory tracking but not updated with obstacle detections, ensuring a fair comparison with baseline methods.
%For the dynamic simulations with the MPC control layer, the VPM was initialized for tracking but not updated with obstacle detections, ensuring a fair comparison with baseline methods. 

\begin{figure}[tb]
\centerline{\includegraphics[width=0.95\columnwidth]{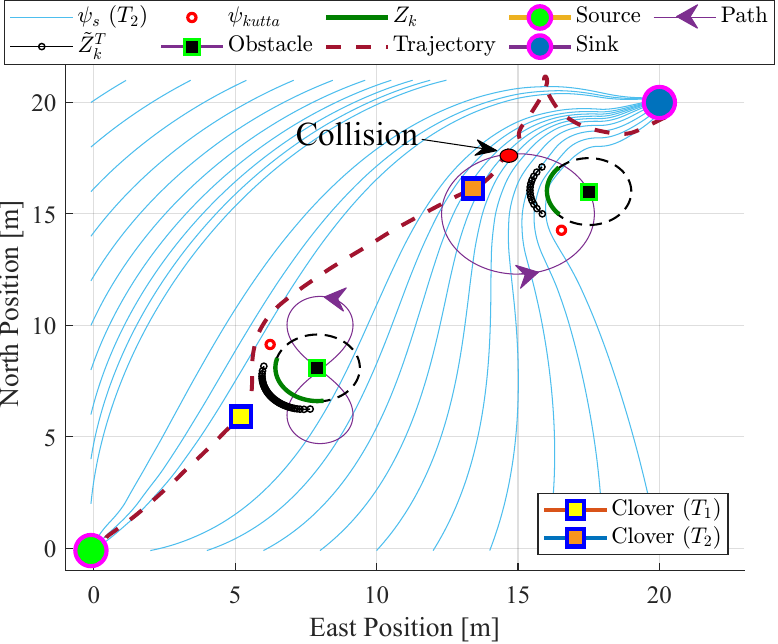}}
\caption{Clover Gazebo simulation showing global trajectory profiles generated by the VPM-A algorithm. The resultant velocity field guides the Clover drone to its destination while avoiding slow moving dynamic cylinders. Two timestamps, $T_1 = 8.9\,\text{s}$ and $T_2=24.74\,\text{s}$, along the trajectory, and a collision at $T_3=26.25\,\text{s}$ are displayed.}
\label{fig:VPM_dynamic}
\end{figure}

\subsubsection{Simple Dynamic}\label{sec:slow_obs}

As an initial dynamic scenario, two cylindrical obstacles with a radius of $1.5\,\text{m}$ are programmed to follow smooth, complex trajectories defined by their position, velocity, and acceleration. 

\paragraph*{\textbf{{Case A)}} Slow-Moving Obstacles}

Obstacle $O_1$, centered at $(8,8)\,\text{m}$, follows a Lemniscate of Bernoulli trajectory with a radius of $3\,\text{m}$, while obstacle $O_2$, centered at $(15,15)\,\text{m}$, follows a circular trajectory with a radius of $2\,\text{m}$. 

The obstacles move at relatively low speeds:
\begin{itemize}
    \item Obstacle $O_1$ completes its trajectory in $28\,\text{s}$, reaching a velocity of $0.8\,\text{m/s}$ and an acceleration of $0.6\,\text{m/s}^2$.
    \item Obstacle $O_2$ completes its trajectory in $17\,\text{s}$, reaching a velocity of $0.98\,\text{m/s}$ and an acceleration of $0.38\,\text{m/s}^2$.
\end{itemize}

\paragraph*{\textbf{{Case B)}} Fast-Moving Obstacles}
A second scenario considers a higher-speed version of the same trajectories:
\begin{itemize}
    \item Obstacle $O_1$ completes its trajectory in $8.5\,\text{s}$, reaching a velocity of $2.8\,\text{m/s}$ and an acceleration of $6.2\,\text{m/s}^2$.
    \item Obstacle $O_2$ completes its trajectory in $7.7\,\text{s}$, reaching a velocity of $1.8\,\text{m/s}$ and an acceleration of $1.5\,\text{m/s}^2$.
\end{itemize}

\begin{table}[bp]
\caption{Collision-free rate for each algorithm, evaluated over 10 separate simulations or experiments.}
\centering
\begin{tabular}{ c|>{\centering}p{1.05cm}|>{\centering}p{1.2cm}|>{\centering}p{0.9cm}|c}
\hline
\multicolumn{5}{c}{Gazebo Simulations}\\
 \hline
 Algorithm & Static & Simple Dynamic [Case A, Case B] & Complex Dynamic & 3D \\
 \hline \hline
 VPM-A & 100\%  & [70,30]\%  & \textemdash  & \textemdash \\
 VPM-B ($\xi = \pm 0.5$) & 100\%  & [60,20]\%  & \textemdash  & \textemdash \\
 VPM-B ($\xi = \pm 0.3$) & 100\%  & [80,40]\%  & \textemdash  & \textemdash \\
 APF & 0\%  & \textemdash  & \textemdash  & \textemdash \\
 \hline
 VPM*-MPC-HOCBF-AKF & [0*,100]\%  & [100,100]\%  & 90\%  & 90\% \\
 VPM*-MPC-AKF & \textemdash  & [100,100]\%  & 90\%  & 70\% \\
 VPM*-MPC & \textemdash  & [100,70]\%  & 40\%  & 30\% \\
 \hline
 \multicolumn{5}{c}{Hardware Experiments}\\
 \hline
 Algorithm & \multicolumn{2}{c|}{Static} & \multicolumn{2}{c}{Static - 3D}\\
 \hline\hline
 VPM-A & \multicolumn{2}{c|}{100\%} & \multicolumn{2}{c}{\textemdash}\\
 VPM-B ($\xi = \pm 0.5$) & \multicolumn{2}{c|}{80\%} & \multicolumn{2}{c}{\textemdash}\\
 VPM-B ($\xi = \pm 0.4$) & \multicolumn{2}{c|}{90\%} & \multicolumn{2}{c}{\textemdash}\\
 VPM-B ($\xi = \pm 0.3$) & \multicolumn{2}{c|}{100\%} & \multicolumn{2}{c}{\textemdash}\\
 VPM{*}-MPC-HOCBF-AKF & \multicolumn{2}{c|}{100\%} & \multicolumn{2}{c}{100\%}\\
 \hline
\end{tabular}
\label{table:MAE}
\end{table}

From Table \ref{table:MAE}, in Case A, the VPM-A achieved collision-free tracking in $70\%$ of the simulations using the safety parameters in Table \ref{table:ctrlparam}, while the VPM-B achieved collision-free tracking in $60-80\%$ of the simulations, influenced by the stream function control parameter $\xi$. An example run where VPM-A failed is shown in Fig. \ref{fig:VPM_dynamic}. Upon detecting $O_1$, LiDAR readings $\bm{Z}_k$ are transformed into $\Tilde{\bm{Z}}^T_k$, and the Kutta condition $\psi_{kutta}$ is extended from the trailing edge to direct fluid flow, keeping the drone at a safe distance. However, when detecting $O_2$, the Clover is on a collision course. Despite VPM-A updating the velocity field and trajectory, the avoidance attempt fails, leading to a destabilizing collision.

%From Table \ref{table:MAE}, in Case A, VPM-A achieved collision-free tracking in $70\%$ of the simulations using the safety parameters in Table \ref{table:ctrlparam}, while VPM-B achieved collision-free tracking in $60-80\%$ of the simulations, depending on the stream function control parameter $\xi$. An example run with VPM-A that resulted in a collision is shown in Fig. \ref{fig:VPM_dynamic}. When the Clover first detects $O_1$, the LiDAR readings $\bm{Z}_k$ are transformed into $\Tilde{\bm{Z}}^T_k$, and the Kutta condition $\psi_{kutta}$ is extended from the trailing edge of the surface to direct fluid flow, enabling the drone to maintain a safe distance from the obstacle. However, upon detecting obstacle $O_2$, the drone is on a collision course. Although VPM-A updates the velocity field and trajectory, the Clover's attempt to avoid the obstacle fails, resulting in a collision that destabilizes the drone. 

While tuning safety parameters and update rates in the quasi-steady VPM improves performance, the algorithm is limited by computational speed and, more critically, control authority, particularly in highly dynamic scenarios. Position and velocity-based control introduce lag, and may require large, sudden velocity changes. These limitations become more pronounced in Case B simulations, where higher obstacle velocities and accelerations further degrade performance (see Table \ref{table:MAE}), highlighting the need for more robust methods. High relative-degree control, such as acceleration with the MPC approach, enables precise maneuvering \cite{xiao2023barriernet} and improves responsiveness to rapidly accelerating obstacles.

\subsubsection{Complex Dynamic}\label{sec: MPC_fast}

In this simulation, two complex-shaped obstacles, each composed of an array of cylinders with a radius $1.5\,\text{m}$ per cylinder, follow smooth trajectories defined by position, velocity, and acceleration:
\begin{itemize}
    \item Obstacle $O_1$ follows a Lemniscate of Bernouli trajectory (center: $(11,10)\,\text{m}$, radius: $4\,\text{m}$), completed in $16\,\text{s}$ reaching a velocity of $1.6\,\text{m/s}$, an acceleration of $1.9\,\text{m/s}^2$, and a maximum rotational rate of $0.5\,\text{rad/s}$.
    \item Obstacle $O_2$ follows a circular trajectory (center: $(30,17)\,\text{m}$, radius: $5\,\text{m}$), completed in $15\,\text{s}$ reaching a velocity of $2.1\,\text{m/s}$, an acceleration of $0.9\,\text{m/s}^2$, and a maximum rotational rate of $0.55\,\text{rad/s}$.   
\end{itemize}

\begin{figure}[t]
\centerline{\includegraphics[width=\columnwidth]{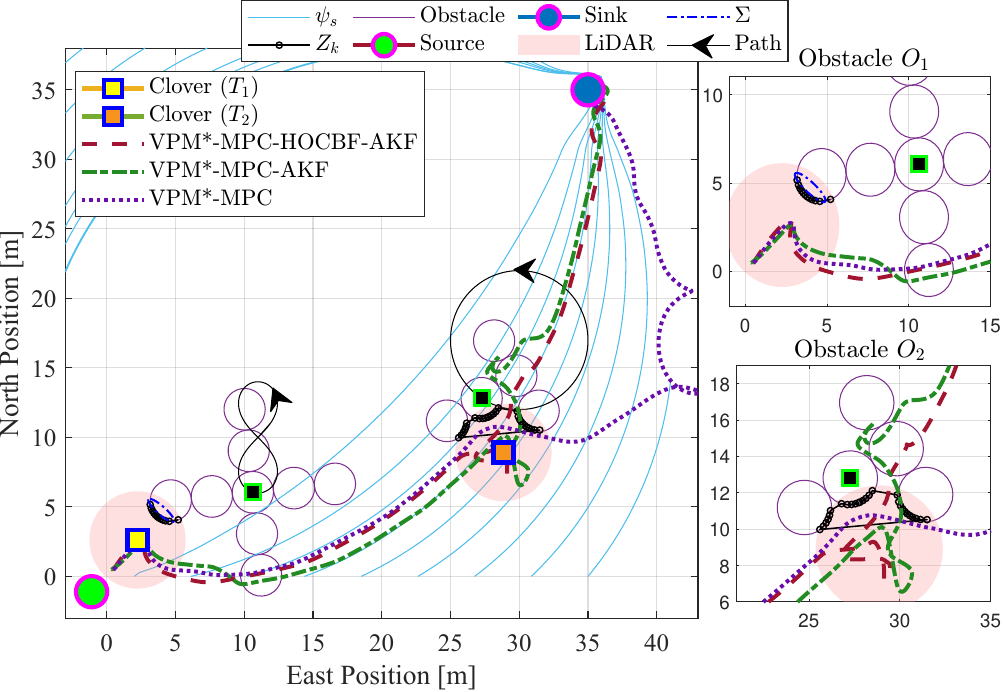}}
\caption{Clover Gazebo simulation illustrating global trajectory profiles and obstacle avoidance using the three different MPC approaches in the presence of rapidly accelerating and rotating obstacles with complex shapes. Two timestamps, $T_1 = 3.23\,\text{s}$ and $T_2=32.84\,\text{s}$, along the trajectory are displayed.}
\label{fig:mpc_dynamic}
\end{figure}

\begin{figure}[h]
\centerline{\includegraphics[width=0.95\columnwidth]{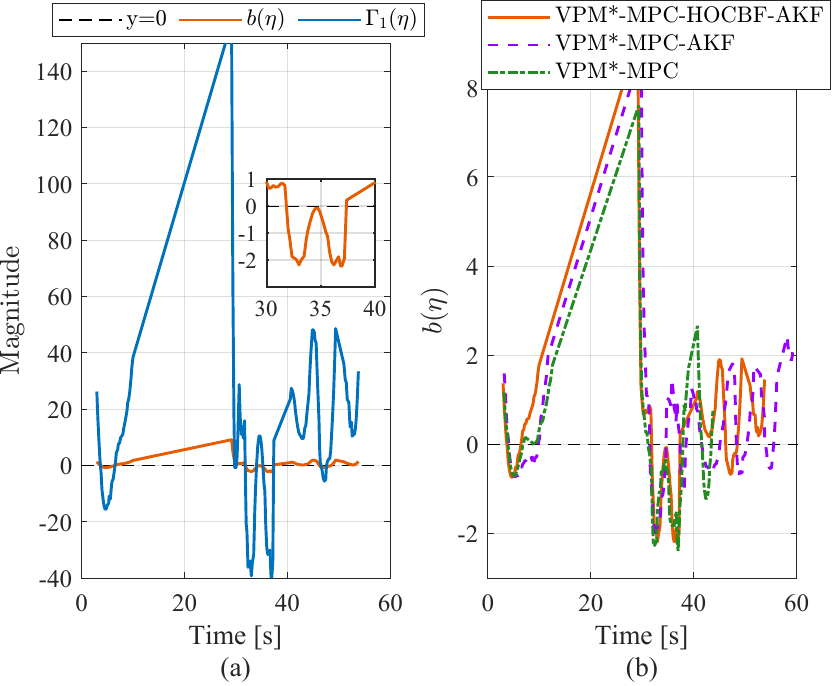}}
\caption{Clover Gazebo simulation showing the evolution profiles of functions $b(\boldsymbol{\eta})$ and $\Gamma(\boldsymbol{\eta})$ for dynamic obstacle avoidance: (a) $b(\boldsymbol{\eta})$ and $\Gamma(\boldsymbol{\eta})$ for the VPM*-MPC-HOCBF-AKF; (b) $b(\boldsymbol{\eta})$ for each of the three MPC approaches. The conditions $b(\boldsymbol{\eta}) \geq 0$ and $\Gamma(\boldsymbol{\eta}) \geq 0$ imply forward invariance of $C_1\cap C_2$.}
\label{fig:b_psi_dynamic}
\end{figure}

From Table \ref{table:MAE}, the VPM*-MPC-HOCBF-AKF algorithm achieved a $90\%$ success rate. A collision-free flight is depicted in Fig. \ref{fig:mpc_dynamic}. Initially on a collision course with $O_1$, the Clover drone adapts its trajectory using the MPC solver, which integrates VPM-generated trajectory tracking, AKF obstacle detection, and HOCBF constraints \eqref{eg:mpc_in_con_2}. The MPC input accelerates the drone away from $O_1$, maintaining a safe distance. As seen in Fig. \ref{fig:b_psi_dynamic}, the forward invariance of sets $C_1\cap C_2$ were mostly preserved, where $C_1 = \{ \boldsymbol{\eta} \in \mathbb{R}  :  b(\boldsymbol{\eta}) \geq 0 \}$ and $C_2 = \{ \boldsymbol{\eta} \in \mathbb{R}  :  \Gamma_{1}(\boldsymbol{\eta}) \geq 0 \}$. Despite the limitations of the linear acceleration model \eqref{eq:linear_acceleration}, the Clover demonstrated high performance, aided by the open-loop dynamics and uncertainty propagation described in Section \ref{sec:uncertainty}.

%Upon detecting $O_1$, the Clover drone was initially on a collision course. However, the MPC solver combined the VPM-generated trajectory tracking with the AKF obstacle detection feedback and the HOCBF constraints \eqref{eg:mpc_in_con_2} to compute a control input that accelerated the Clover away from the incoming obstacle, maintaining a safe distance from $O_1$. As seen in Fig. \ref{fig:b_psi_dynamic}, the forward invariance of sets $C_1\cap C_2$ were mostly preserved, where $C_1 = \{ \boldsymbol{\eta} \in \mathbb{R}  :  b(\boldsymbol{\eta}) \geq 0 \}$ and $C_2 = \{ \boldsymbol{\eta} \in \mathbb{R}  :  \Gamma_{1}(\boldsymbol{\eta}) \geq 0 \}$. Despite the limited accuracy of the linear acceleration model \eqref{eq:linear_acceleration}, the Clover was still able to exhibit high performance, which was aided by the open-loop dynamics and uncertainty propagation described in Section \ref{sec:uncertainty}. 

Similarly, the VPM*-MPC-AKF algorithm ensures effective obstacle avoidance by leveraging AKF-based obstacle state prediction, accounting for uncertainties in LiDAR feedback and obstacle state estimation. The Euclidean norm geometric constraint $b \geq r$ is maintained within the prediction horizon. The proposed VPM*-MPC-HOCBF-AKF optimizes safety, speed, and efficiency, ensuring low speed variance, minimal control effort, and sufficient obstacle clearance, outperforming other methods as shown in Table \ref{table:mpc_dynamic}.

%Similarly, the VPM*-MPC-AKF algorithm demonstrated effective obstacle avoidance by leveraging the proposed AKF for obstacle state prediction, which accounts for uncertainties in LiDAR feedback and pose estimations. The Euclidean norm geometric constraint $b \geq r$ was managed within the prediction horizon. Considering the higher-order dynamics, the proposed VPM*-MPC-HOCBF-AKF balances safety constraints with speed and efficiency by ensuring low speed variance, minimal control effort, and sufficient distance from the obstacle, compared to other methods, as shown in Table \ref{table:mpc_dynamic}.

In contrast, the standard MPC algorithm with the Euclidean norm constraint performs poorly. The Clover struggles to predict and react to dynamic obstacles, only attempting avoidance when the trajectories approach the constraint boundary $b \approx 0$. This results in abrupt trajectory shifts, higher speed variance, and increased collision rates.

\subsection{Extension to 3D}
In Sections \ref{sec:MPC_static} and \ref{sec: MPC_fast}, the VPM*-MPC-HOCBF-AKF is used as a reactive obstacle avoidance strategy, leveraging 2D LiDAR measurements and AKF-based state estimation. Since 3D sensor-based obstacle state estimation is beyond the scope of this paper, this simulation assumes that the current obstacle center position $\bm{p}_{O}^i$ is known and used as observations for the AKF. Future states within the prediction horizon $T$ at $N$ shooting nodes are estimated using the open-loop forward model described in Section \ref{sec:uncertainty}.

For this simulation, an obstacle-free velocity field is generated using the VPM, providing a trajectory for the Clover to track while maintaining a fixed altitude of $3.8\,\text{m}$. Two spherical obstacles, each with a radius of $1.5\,\text{m}$ are programmed to follow smooth, complex trajectories:
\begin{itemize}
    \item Obstacle $O_1$ follows a Torus trajectory (major radius: $3.0\,\text{m}$, minor radius: $1.5\,\text{m}$), completed in $4.2\,\text{s}$ reaching a velocity of $7.1\,\text{m/s}$ and an acceleration of $13.6\,\text{m/s}^2$.
    \item Obstacle $O_2$ follows a Lissajous trajectory (amplitude: $2.5\,\text{m}$ along each axis), completed in $9.0\,\text{s}$ reaching a velocity of $7.9\,\text{m/s}$ and an acceleration of $19.8\,\text{m/s}^2$.
\end{itemize}
The high-speed trajectories provide a robust test for the avoidance algorithms.

%For this simulation, an obstacle-free velocity field was generated using the VPM, providing a trajectory for the Clover to track while maintaining a fixed altitude of $3.8,m$. Two spherical obstacles, each with a radius of $1.5\,m$ were programmed to follow smooth, complex trajectories. Obstacle $O_1$ followed a torus trajectory with a major radius of $3.0\,m$ and a minor radius of $1.5\,m$ while obstacle $O_2$ followed a Lissajous curve with an amplitude of $2.5\,m$ along each axis $(x, y, z)$, defined by sinusoidal motion. 
%\begin{itemize}
   % \item Obstacle $O_1$: Completed the torus trajectory in $4.2\,s$ reaching a velocity of $7.1\, m/s$ and an acceleration of $13.6\,m/s^2$.
   % \item Obstacle $O_2$: Completed the Lissajous curve in $9.0\,s$ reaching a velocity of $7.9\,m/s$ and an acceleration of $19.8\,m/s^2$.
%\end{itemize}
%The high-speed trajectories provided a robust test for the avoidance algorithms.

\begin{figure}[tb]
\centerline{\includegraphics[width=\columnwidth]{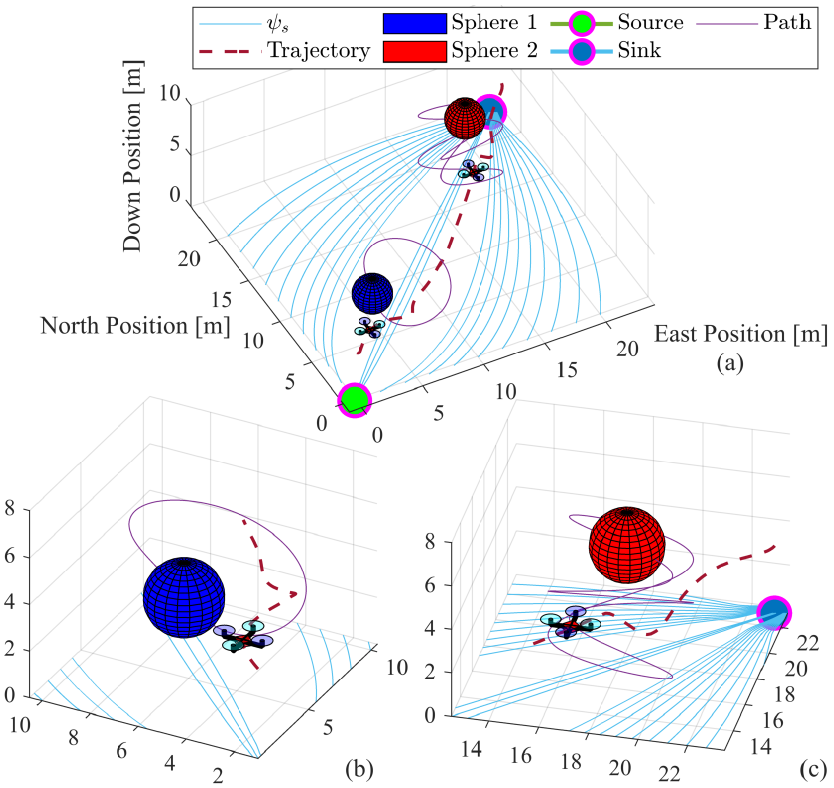}}
\caption{Clover Gazebo simulation showing global 3D trajectory profiles and obstacle avoidance generated by the VPM*-MPC-HOCBF-AKF algorithm in the presence of rapidly accelerating dynamic spheres. Two timestamps, $T_1 = 3.7\,\text{s}$ and $T_2=23.7\,\text{s}$, along the trajectory are displayed.}
\label{fig:mpc_3D}
\end{figure}

\begin{figure}[tb]
\centerline{\includegraphics[width=\columnwidth]{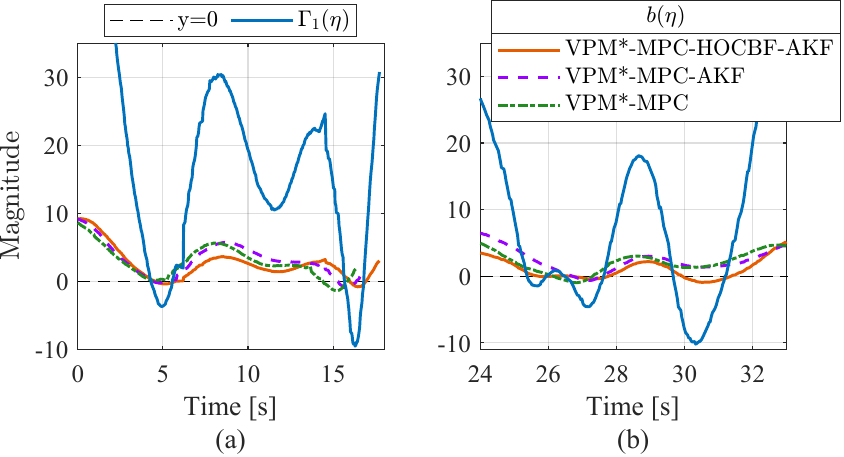}}
\caption{Clover Gazebo simulation showing evolution profiles of functions $b(\boldsymbol{\eta})$ and $\Gamma(\boldsymbol{\eta})$ for 3D dynamic obstacle avoidance: (a) Avoiding obstacle $O_1$; (b) Avoiding obstacle $O_2$. $b(\boldsymbol{\eta}) \geq 0$ and $\Gamma(\boldsymbol{\eta}) \geq 0$ imply forward invariance of $C_1\cap C_2$.}
\label{fig:b_psi_3D}
\end{figure}

As shown in Table \ref{table:MAE}, the VPM*-MPC-HOCBF-AKF achieved a $90\%$ collision-free rate across ten simulations. Unlike in the 2D cases, the HOCBF constraint \eqref{eg:mpc_in_con_2} was not limited by the LiDAR sensor’s range. Due to the high velocities of the spheres, the $(\beta_1, \beta_2)$ values (see Table \ref{table:ctrlparam}) were adjusted to activate the HOCBF earlier, allowing the Clover to make corrections further in advance. The parameter adjustment reduces the likelihood of unnecessarily large control inputs and conflicts with constraints \eqref{eq:mpc_in_con_1}. Conversely, the VPM*-MPC-AKF and VPM*-MPC exhibit lower success rates (see Table \ref{table:MAE}) and higher velocity variances (see Table \ref{table:mpc_dynamic}), indicating delayed reactions due to the nature of the Euclidean norm constraint. Without the AKF for obstacle state prediction, the MPC-based method struggles to react effectively to the high-acceleration spheres, leading to frequent collisions.

%The main cause of the single collision was the severity of the relative speed and acceleration between the obstacles and the Clover, combined with a relatively small safety distance and the limitations of the Clover system during the simulation. Additionally, the linear dynamic modelling (??) expects quicker reactions to control input from the ACADOS solver, however, applying to a Gazebo physics system exemplifying increased realism, this is not the case. 

A successful flight is illustrated in Fig. \ref{fig:mpc_3D}, with $b(\boldsymbol{\eta})$ and $\Gamma(\boldsymbol{\eta})$ depicted in Fig. \ref{fig:b_psi_3D}. From Fig. \ref{fig:mpc_3D}(b), Sphere $O_1$ approaches from the left, prompting the Clover to slow down and take an overhead path to avoid collision. This close encounter occurs at $t = 25\,\text{s}$, where Fig. \ref{fig:b_psi_3D}(a) shows $b \leq r$ for $t \in [24, 25]\,\text{s}$. Shortly after, at $t=35\,\text{s}$, the Clover accelerates to avoid a rear approach as Sphere $O_1$ loops back. The high-acceleration Lissajous trajectory of Sphere $O_2$ posed additional challenges, but the MPC solver effectively generated acceleration setpoints, enabling the Clover to quickly adjust and avoid Sphere $O_2$.

\subsection{Computational Resources}
All the simulations are hosted on a machine with an Intel\textsuperscript{\textregistered} Core\texttrademark\ i7-10750H CPU (four cores, $2.6\,\text{GHz}$ base frequency), $8\,\text{GB}$ of RAM, and a Linux-based operating system. The VPM computations are executed quasi-steadily at $0.8$–$1.5\,\text{Hz}$ when obstacles are detected, with an average iteration time of $0.37 \,\text{s}$ (range: $0.1$ to $0.7 \,\text{s}$). Additionally, CPU usage averages $47\%$, peaking to $70$–$90\%$, highlighting the need for a dedicated processor to ensure reliable real-time performance.

%NOTE!: Unlike in the original HOCBF ref which uses a single step otptimal controller with HOCBF, the MPC-HOCBF can use the future predictions to avoid current violations of the HOCBF conditions (does not activate as the activation in the future is avoided to some extent, ideally).

%While the MPC-HOCBF was used as a more reactive algorithm in the 2D case, due to 3D obstacle state estimation being out of the scope of this paper, we will assume the position in 3D space is known at an instant. As a result, the MPC-HOCBF can take advantage of this, and provide re-planning with more time in advance. This will outline the true benefits of this algorithm with its ability to predict future states and react to dynamic obstacles.

\section{Experimental Results}
%This section presents experimental results from a multi-phase experiment designed to replicate the simulation tests with comparable conditions, demonstrating the effectiveness of the proposed control algorithms.

%This section presents experimental results from a multi-phase study replicating simulation conditions to evaluate the effectiveness of the proposed control algorithms.

This section presents experimental results from a multi-phase experiment designed to replicate the simulation tests under comparable conditions, demonstrating the effectiveness of the proposed control algorithms.

%For the Gazebo simulation environment used in Section \ref{sec:simulations}, simulations were hosted on a machine configured with four cores of an Intel\textsuperscript{\textregistered} Core\texttrademark\ i7-10750H CPU, each operating at a base frequency of $2.6\,\text{GHz}$. The machine was allocated $8\,\text{GB}$ of RAM and ran on a Linux-based operating system. During simulations, the VPM computations were performed in a quasi-steady manner, executing at a designated frequency between $0.8-1.5\,\text{Hz}$ when obstacles were detected. The average execution time per iteration was $0.37 \,s$ (ranging from $0.1$ to $0.7 \,s$). Additionally, CPU usage averaged $47\%$ with jumps to $70-90\%$, highlighting the need for a dedicated processor to ensure reliable real-time performance.

\begin{figure}[tb]
\centerline{\includegraphics[width=0.75\columnwidth]{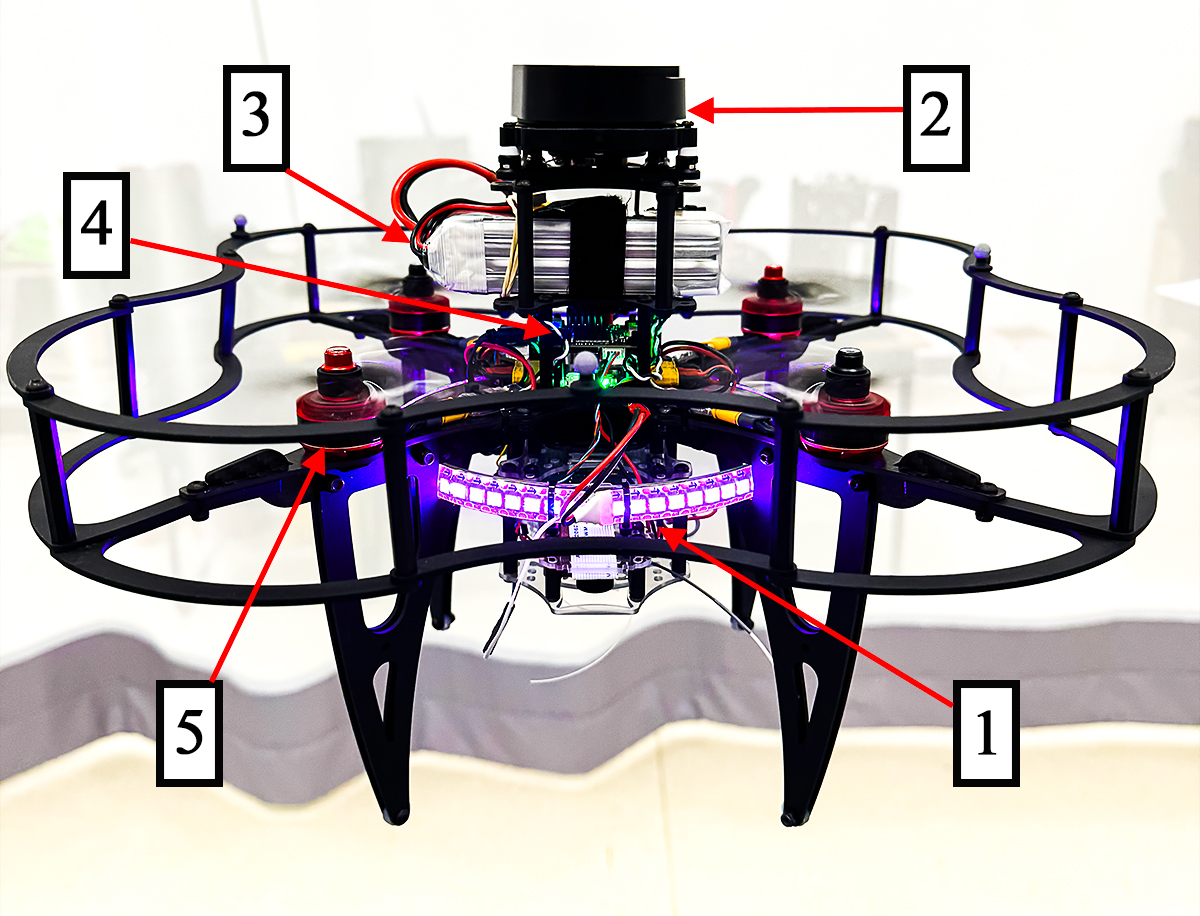}}
\caption{COEX Clover 4.2 platform: 1. Raspberry Pi 4 Model B, 2. Laser Distance Sensor LDS-01, 3. Tattu 2300 mAh 4S 45C LiPo battery, 4. COEX Pix flight controller, 5. COEX BR2306 2400-kV motors.}
\label{fig:clover_flight}
\end{figure}

The experiments are carried out on a COEX Clover quadrotor, shown in Fig. \ref{fig:clover_flight}, which features a COEX Pix flight controller running PX4 firmware, with attitude controllers tuned using PX4’s adaptive auto-tune module for reliable quaternion tracking. The onboard Raspberry Pi 4B, running ROS with the Clover image, communicates with the ground station (QGroundControl) and the COEX Pix flight controller via MAVROS, facilitating the execution of high-level control algorithms. Additionally, a 360$^{\circ}$ LDS-01 LiDAR provides real-time laser-based navigation.

%featuring labeled components, is shown in Fig. \ref{fig:clover_flight}. The COEX Pix flight controller, running the PX4 firmware was utilized. The attitude controllers were tuned using the adaptive auto-tune control module provided by PX4. This allows for the assumption of sufficient tracking of the quaternion setpoints. A Raspberry Pi 4B onboard computer running ROS, equipped with the Clover image, communicates with both the ground station computer running QGroundControl and the COEX Pix flight controller. This setup facilitates the execution of high-level control algorithms through MAVROS. Additionally, a 360$^{\circ}$ LDS-01 was attached to provide laser-based navigation.

%The COEX Clover quadrotor, featuring labeled components, is shown in Fig. \ref{fig:clover_flight}. The COEX Pix flight controller, running the PX4 firmware was utilized. The attitude controllers were tuned using the adaptive auto-tune control module provided by PX4. This allows for the assumption of sufficient tracking of the quaternion setpoints. A Raspberry Pi 4B onboard computer running ROS, equipped with the Clover image, communicates with both the ground station computer running QGroundControl and the COEX Pix flight controller. This setup facilitates the execution of high-level control algorithms through MAVROS. Additionally, a 360$^{\circ}$ LDS-01 was attached to provide laser-based navigation.
Due to the limited indoor experimental area, dynamic test cases could not be performed safely and were therefore left for simulation. Experiments on static obstacles were carried out as follows.

\begin{figure}[tb]
\centerline{\includegraphics[width=0.95\columnwidth]{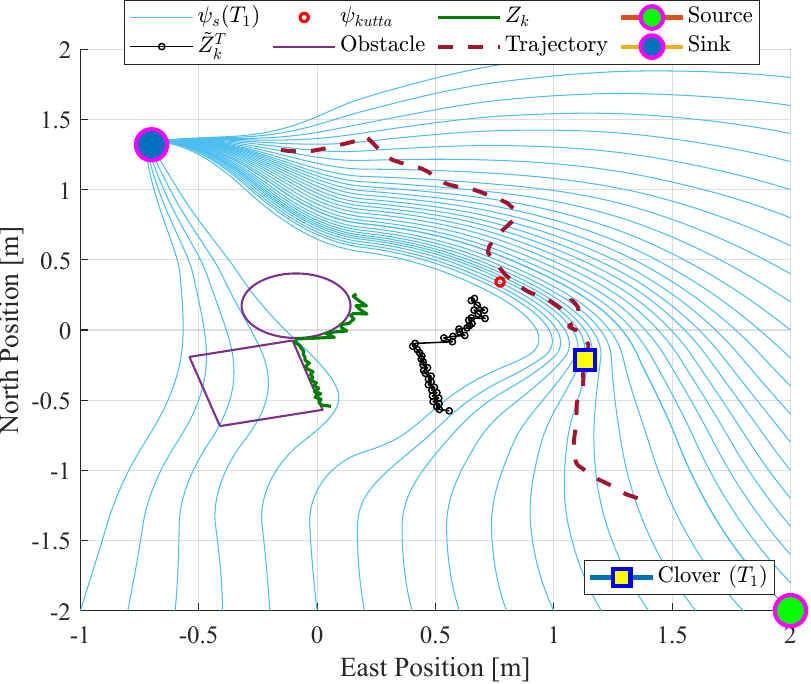}}
\caption{Hardware experiment showcasing the global trajectory generated by the VPM-A algorithm. The resultant velocity field guides the Clover drone to its destination while avoiding static obstacles. A timestamp of $T_1=4.6\,\text{s}$ is displayed.}
\label{fig:vpm_hardware}
\end{figure}

\subsection{COEX Clover LiDAR Navigation with Static Obstacle}

\subsubsection{Online Vortex Panel Method}

In the static obstacle avoidance scenario illustrated in Fig. \ref{fig:vpm_hardware}, the Clover is tasked with navigating from a designated takeoff point to a proximity of a sink within the motion capture system's volume, while avoiding an irregular concave obstacle. This is achieved by tracking a velocity field generated by the VPM-A. Similar to the Gazebo simulations in Section \ref{sec:VPM_gaz_static}, the algorithm constructs a virtual 2D rigid surface from the detected irregular-shaped surface. The virtual rigid surface $\Tilde{\bm{Z}}_k^T$, combined with the Kutta condition $\psi_{kutta}$, directs the vortex flow around the obstacle, enabling the Clover to navigate without any collisions.

%The precision of the lidar...
 A noticeable effect not observed in the simulation is the wall effect. When the Clover approaches the obstacles at close proximity, the wall effect induces a moment that tilts the Clover towards the obstacles. As seen in Fig. \ref{fig:clover_flight}, the addition of the LDS makes the Clover top-heavy, increasing the difficulty of maintaining stable attitude control. These disturbances are more pronounced within the confined space of the motion capture system volume, underscoring the need for robust position control \cite{Smith2024}.

 \begin{figure}[tb]
    \centering
    \subfloat[]{
        \includegraphics[width=0.95\columnwidth]{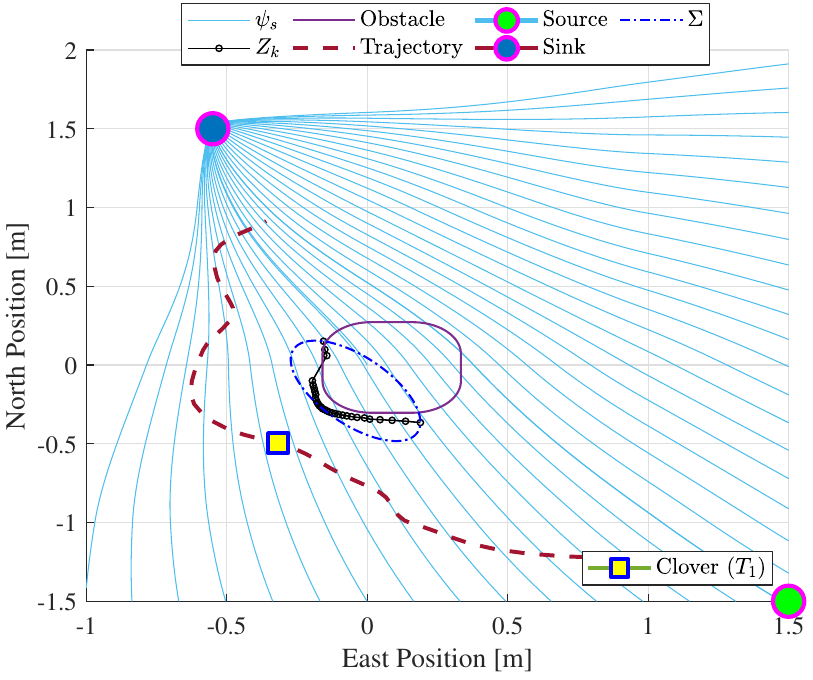}
        \label{fig:mpc_2D_exp}
    }\\ % Add a line break to stack the subfigures
    \subfloat[]{
        \includegraphics[width=0.7\columnwidth]{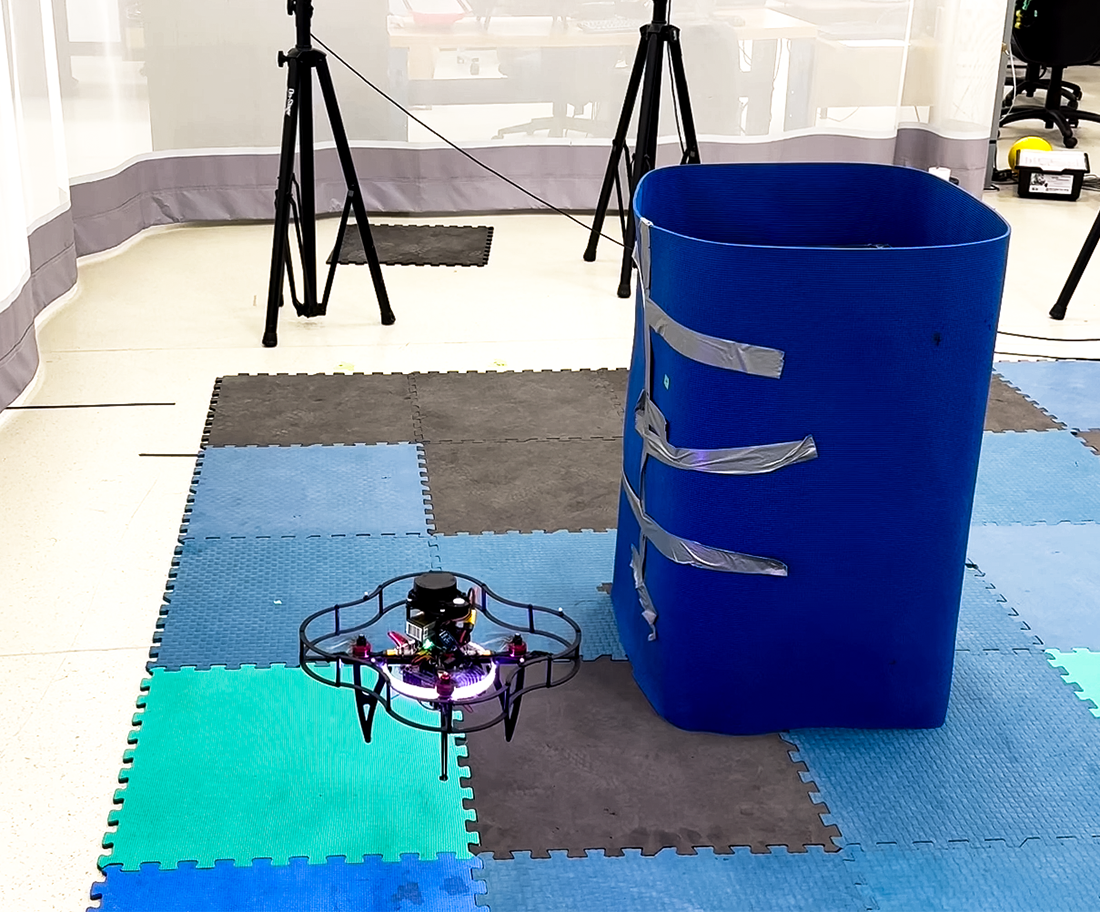}
    }
    \caption{(a) Hardware experiment demonstrating the global trajectory and obstacle avoidance achieved by the VPM*-MPC-HOCBF-AKF algorithm with a static obstacle. A timestamp of $T_1=4.1\,\text{s}$ is displayed. (b) Layout of the hardware setup used in the experiment.}
    \label{fig:mpc_2D_hard}
\end{figure}

\subsubsection{VPM-MPC-HOCBF-AKF}

Similar to the VPM method discussed in the previous section, the Clover is tasked with navigating around an obstacle using the VPM*-MPC-HOCBF-AKF approach (see Fig. \ref{fig:mpc_2D_hard}). The initialized VPM-generated field provides setpoints for the MPC, while the AKF uses MBE observations to supply feedback to the HOCBF, managing the relative position between the Clover and the obstacle. This integration enables the Clover to safely navigate within the motion capture system volume and reach the sink location, despite significant aerodynamic disturbances and wall effects.

% \begin{figure}[hbt]
% \centerline{\includegraphics[width=0.9\columnwidth]{images/mpc_2D_exp_2.pdf}}
% \caption{Hardware experiment showcasing the global trajectory and obstacle avoidance generated by the VPM-MPC-HOCBF algorithm while avoiding a static obstacle.}
% \label{fig:mpc_2D_exp}
% \end{figure}

The parameters $(\beta_1, \beta_2)$ (see Table \ref{table:ctrlparam}) are selected to be sufficiently low to prevent late activation of the HOCBF. Setting these parameters too high would cause the Clover to navigate close to the obstacle before accelerating sharply to reach more compliant sets $C_1 \cap C_2$, which is undesirable for a small experimental area. 

%Additionally, without the introduction of slack variables, violation of the HOCBF hard set constraints would result in SQP-RTI errors from the ACADOS solver, causing the Clover to autoland.

%needed to be reduced compared to the Gazebo simulations (see Table \ref{table:ctrlparam}). This adjustment was necessary due to the closer proximity to obstacles during the actual experiments compared to the simulations. If $(\beta_1, \beta_2)$ were set too large, the sets $C_1 \cap C_2$ could be violated upon takeoff. Depending on the strictness of the constraint slack variables, this scenario may lead to the Clover accelerating into more compliant sets $C_1 \cap C_2$ or result in SQP-RTI errors from the ACADOS solver, leading the Clover to autoland.

% In the document body
\begin{figure}[tb]
    \centering
    \subfloat[]{
        \includegraphics[width=0.95\columnwidth]{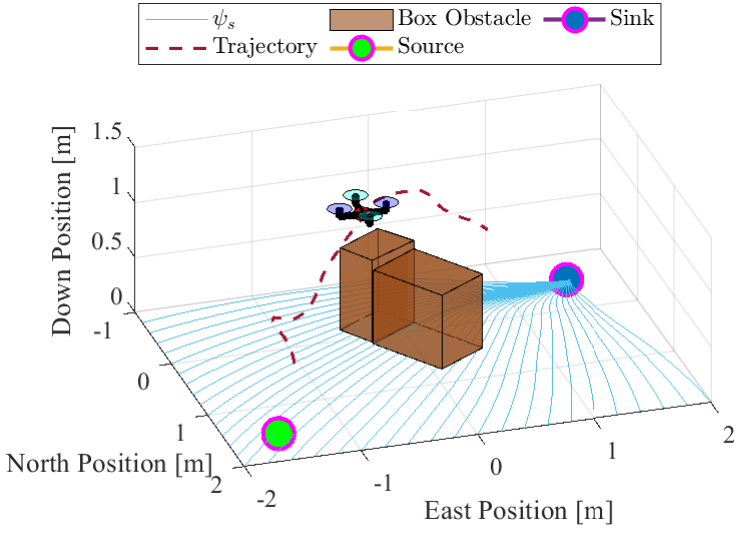}
        \label{fig:mpc_3D_exp}
    }\\ % Add a line break to stack the subfigures
    \subfloat[]{
        \includegraphics[width=0.75\columnwidth]{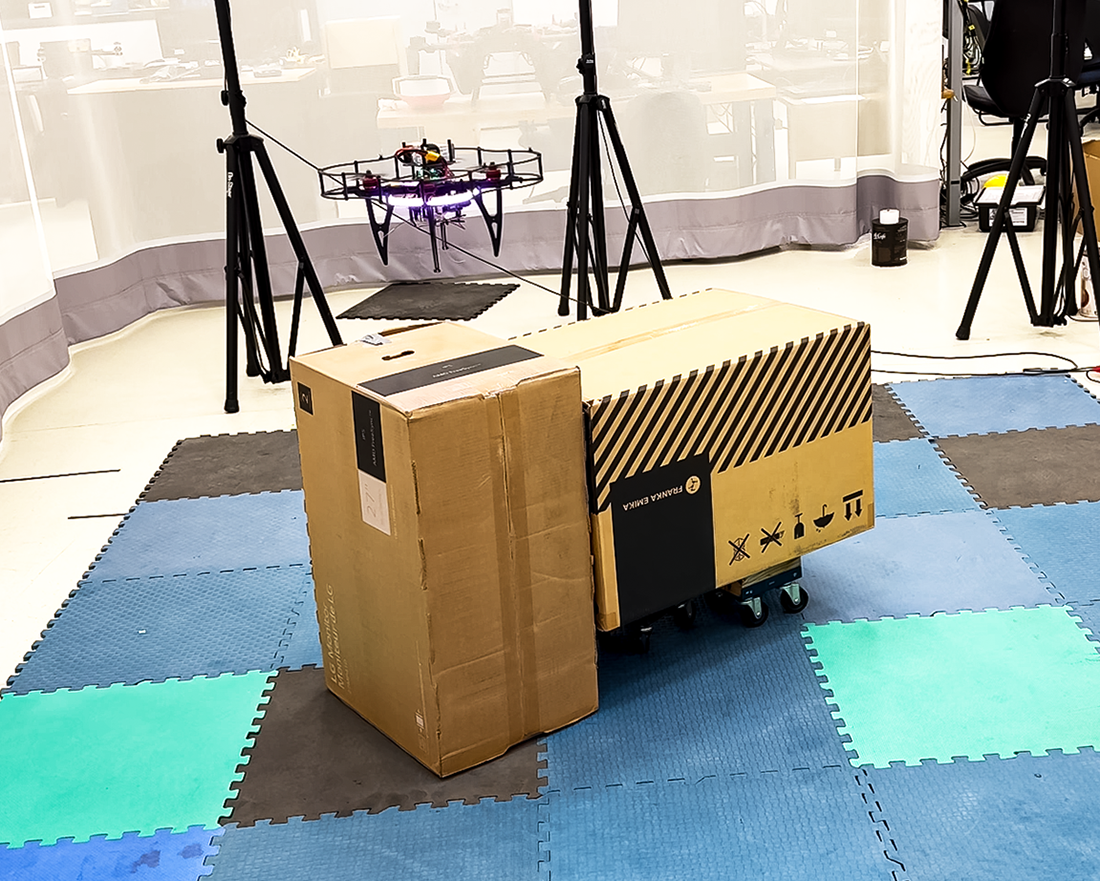}
        \label{fig:subfigure_example}
    }
    \caption{(a) Hardware experiment demonstrating the global 3D trajectory and obstacle avoidance achieved by the VPM*-MPC-HOCBF algorithm while avoiding a static boxed structure. (b) Layout of the hardware setup used in the experiment.}
    \label{fig:mpc_combined}
\end{figure}

\subsection{Static 3D}

The VPM*-MPC-HOCBF-AKF method is extended to a 3D scenario in an experiment designed to validate the algorithm in such cases. The obstacle consists of two tall boxes, as illustrated in Fig. \ref{fig:mpc_combined}, with the center approximated using the motion capture system for the MPC-HOCBF solver. To prevent the Clover from tracking in-plane and around the obstacle, the weights $\bm{W}$ on the constant-altitude position setpoint tracking are reduced. These adjustments allowed the Clover to follow an optimal path over the obstacle and land safely at the sink location.

% \begin{figure}[hbt]
% \centerline{\includegraphics[width=0.99\columnwidth]{images/mpc_3d_exp_track.pdf}}
% \caption{Hardware experiment showcasing the global 3D trajectory and obstacle avoidance generated by the VPM-MPC-HOCBF algorithm while avoiding a static boxed structure.}
% \label{fig:mpc_3D_exp}
% \end{figure}

% \begin{figure}[hbt]
% \centerline{\includegraphics[width=0.99\columnwidth]{images/overbox_3D_test.pdf}}
% \caption{Hardware experiment showcasing the global 3D trajectory and obstacle avoidance generated by the VPM-MPC-HOCBF algorithm while avoiding a static boxed structure.}
% \label{fig:yo}
% \end{figure}

\subsection{Discussion}
Onboard the Clover is a Raspberry Pi 4B (quad-core Cortex-A72, $1.8 \,\text{GHz}$), which lacks sufficient computational resources to run the VPM alongside existing programs. The VPM, a numerical method for approximating the stream function \eqref{eq:vorticity_distribution}, requires significant processing power. Increasing the LiDAR resolution and the number of panels brings the solution closer to the analytical solution, but at a computational cost. Reducing LiDAR sampling frequency or resolution lowers this demand but at the expense of precision. A more optimal approach would be to offload intensive computations to dedicated hardware, such as a Field-Programmable Gate Array (FPGA) or a GPU-based processor. However, due to hardware constraints, the VPM computations are instead performed offboard on a ground station computer.

\section{Conclusion}

In this article, we proposed a novel VPM fluid flow-based navigation algorithm for quadcopter obstacle avoidance, leveraging 2D inviscid flow dynamics for rigid sail analysis to handle obstacles of arbitrary shapes. The VPM was integrated with an MPC-HOCBF framework, where obstacles were parametrized as MBEs, and an AKF predicted obstacle trajectories, incorporating the relative dynamics of moving obstacles to improve performance. Simulations and experiments showed that the VPM algorithm enables safe navigation around static and slowly moving obstacles of arbitrary shapes, while the MPC-HOCBF-AKF combination improves performance, particularly near accelerating obstacles in close proximity. Future research could extend the proposed developments to 3D sensor-based methods, automate parameter selection, and incorporate predictive avoidance of hybrid systems. 

\section{Acknowledgment}

The authors extend their thanks to Oleg Kalachev from COEX for assistance with the hardware.

\ifCLASSOPTIONcaptionsoff
  \newpage
\fi



%\bibliographystyle{IEEEtran}
\bibliographystyle{IEEEtran}
\bibliography{Quadcopter}

\end{document}